\documentclass[runningheads]{llncs}

\usepackage[T1]{fontenc}

\usepackage{graphicx,tabularx,booktabs}
\usepackage{xurl}
\usepackage{multirow}
\usepackage{tikz}
\usepackage{enumitem}
\usepackage{hyperref}
\usepackage{fontawesome}
\usepackage{makecell}
\usepackage{pgfplots}
\usepackage{circledsteps}
\usepackage{colortbl}
\usetikzlibrary{patterns}
\usepackage[ruled]{algorithm2e}

\pgfplotsset{compat=1.18}
\usepgfplotslibrary{groupplots,fillbetween,statistics}
\usetikzlibrary{external,decorations.pathreplacing,calligraphy,shapes,arrows,fit,positioning}

\usepackage[utf8]{inputenc}
\usepackage{subcaption}
\usepackage{amsmath}
\usepackage{pgfplotstable}
\usetikzlibrary{patterns}
\usetikzlibrary{spy,backgrounds}
\tikzset{external/only named=true}
\usepackage{xstring}
\usepackage{cleveref}
\usepackage{amsfonts}
\usepackage{nicefrac}
\usepackage{microtype}

\definecolor{light}{rgb}{0.68, 0.90, 0.77}
\definecolor{orange}{rgb}{0.93, 0.74, 0.60}
\definecolor{lightorange}{rgb}{1, 0.87, 0.68}
\definecolor{lightgreen}{rgb}{0.76, 0.88, 0.76}
\definecolor{lightgray}{rgb}{0.92, 0.92, 0.92}
\definecolor{lightred}{rgb}{0.92, 0.29, 0.36}
\definecolor{lightpurple}{rgb}{0.92, 0.88, 1}
\definecolor{lightcyan}{rgb}{0.424, 0.651, 0.804}

\tikzset{
 oblique/.style args={#1/#2/#3/#4}{
 condition, 
 minimum width = 2cm,
 minimum height = 2cm,
 below #4 = of #1,
 name=#2,
 append after command={
 node[label={\(\mathbf{w}_{#3}^T\mathbf{x} < b_{#3}\)}, inner sep=0pt, below=8pt] at (\tikzlastnode.center) {}
 }
 }
}

\tikzset{condition/.style={draw, diamond, thick, text centered, minimum height=0.5cm, minimum width=1cm}}
\tikzset{leaf/.style={draw, rectangle, thick, text centered, minimum height=0.5cm, minimum width=1cm}}
\tikzset{line/.style={draw, thick, -latex'}}

\newcommand{\linewitherrordifferent}[5]{
 \addplot [each nth point=1,name path=minuserror,draw=none,no markers,forget plot] table [x={#2},y expr=\thisrow{#3}-\thisrow{#4}] {#1};
 \addplot [each nth point=1,name path=pluserror,draw=none,no markers,forget plot] table [x={#2},y expr=\thisrow{#3}+\thisrow{#4}] {#1};
 \addplot [forget plot,fill=#5,opacity=0.2] fill between[on layer={},of=pluserror and minuserror];
 \addplot [each nth point=1,#5,thick,no markers,legend image code/.code={\fill [fill=#5, opacity=0.2, draw=none] (0mm,-1ex) -- (0mm,1ex) -- (6mm,1ex) -- (6mm,-1ex) -- cycle; \draw [#5,thick] (0mm,0mm) -- (6mm,0mm);}] table [x={#2},y={#3}] {#1};
}

\newsavebox{\mytree}
\savebox{\mytree}{
\begin{tikzpicture}[>=stealth, node distance=2cm, font=\sffamily\small]
 \node[draw, diamond] (root) {};
 \node[draw, rectangle, below left=0.5cm of root] (T) {};
 \node[draw, rectangle, below right=0.5cm of root] (F) {};
 \draw[->] (root) node[below left=0.1cm] {T} -| (T);
 \draw[->] (root) node[below right=0.1cm] {F} -| (F);
\end{tikzpicture}
}

\usetikzlibrary{spy,backgrounds}
\tikzset{external/only named=true}
\crefname{algocf}{alg.}{algs.}
\Crefname{algocf}{Algorithm}{Algorithms}

\definecolor{cbblue}{rgb}{0, 0.447, 0.698}
\definecolor{cbgreen}{rgb}{0, 0.62, 0.45}
\definecolor{cborange}{rgb}{0.84, 0.37, 0}

\begin{document}

\title{Social Interpretable Reinforcement Learning}
\titlerunning{Social Interpretable Reinforcement Learning}

\author{
Leonardo Lucio Custode\orcidID{0000-0002-1652-1690}
\and
Giovanni Iacca\orcidID{0000-0001-9723-1830}
}
\institute{
Department of Information Engineering and Computer Science\\
University of Trento, Italy \\
\email{leonardo.custode@unitn.it}\\
\email{giovanni.iacca@unitn.it}
}

\maketitle 

\begin{abstract}
Reinforcement Learning (RL) bears the promise of being a game-changer in many applications. However, since most of the literature in the field is currently focused on opaque models, the use of RL in high-stakes scenarios, where interpretability is crucial, is still limited. Recently, some approaches to interpretable RL, e.g., based on Decision Trees, have been proposed, but one of the main limitations of these techniques is their training cost. To overcome this limitation, we propose a new method, called Social Interpretable RL (SIRL), that can substantially reduce the number of episodes needed for training. Our method mimics a social learning process, where each agent in a group learns to solve a given task based both on its own individual experience as well as the experience acquired together with its peers. Our approach is divided into the following two phases. (1) In the \emph{collaborative phase}, all the agents in the population interact with a shared instance of the environment, where each agent observes the state and independently proposes an action. Then, voting is performed to choose the action that will actually be deployed in the environment. (2) In the \emph{individual phase}, then, each agent refines its individual performance by interacting with its own instance of the environment. This mechanism makes the agents experience a larger number of episodes with little impact on the computational cost of the process. Our results (on $6$ widely-known RL benchmarks) show that SIRL not only reduces the computational cost by a factor varying from a minimum of 43\% to a maximum 76\%, but it also increases the convergence speed and, often, improves the quality of the solutions.
\end{abstract}


\section{Introduction}
\label{sec:intro}
Interpretable Reinforcement Learning (IRL) is currently considered one of the grand challenges in the field of Interpretable AI \cite{rudin_interpretable_2021}.
In fact, since IRL allows the inspection of the RL models and their (at least partial) understanding, it has the potential to enable a variety of applications of RL that are not possible today. Specifically, interpretability would facilitate the application of RL to real-world scenarios where the learned policies should be translated into actionable regulations, protocols, or laws, e.g., in pandemic control \cite{kompella_reinforcement_2020,custode2022pandemics} or taxation \cite{trott2021building,zheng2022ai}.
In fact, traditional opaque RL methods (based on deep RL) do not allow for a thorough understanding of the models, impeding their validation by domain experts.
However, only recently IRL methods are starting to catch up with deep RL counterparts.

For instance, some approaches for IRL using Decision Trees (DTs) have been proposed in \cite{custode2023evolutionary,custode2021co,custode2022interpretable,dhebar_interpretable-ai_2020,silva_optimization_2020}.
While these approaches seem promising, they have limitations, the most important being the computational cost of the training.
In fact, most of these approaches are based on evolutionary 
RL \cite{custode2023evolutionary,custode2021co,custode2022interpretable},
where, at each step, the optimizer proposes a possibly large set of candidate solutions that must interact with the environment to learn how to solve the task.
This means that, even when using a very small number of episodes for each candidate solution, the total number of interactions with the environment can become very large, thus requiring a significant amount of time (up to several hours, as reported in \cite{custode2021co}).
This makes these approaches unpractical, thus impeding their adoption, especially in computationally expensive environments.

In this work, we borrow ideas from the field of collaborative RL \cite{peng2020non,khadka2019collaborative} to reduce the computational footprint of the interpretable RL methodology proposed in \cite{custode2023evolutionary}. In particular, we improve the efficiency of the baseline methodology by performing partial training in which all the individuals are treated as an ensemble, significantly reducing the link between the number of individuals used in the population-based optimizer and the total number of episodes used for training.
We call this approach \emph{Social Interpretable RL} (SIRL).
Its core concept is inspired by the behavior of social animals and humans, i.e., their capability of learning together how to solve a task. 
To explain SIRL with a simple analogy, consider a class of students who must learn a subject by working on a project that requires some resources, e.g., hardware.
While the simplest thing the teacher could do would be to use one piece of hardware for each student, this would be neither the most efficient (in terms of cost) nor the most effective (in terms of learning outcome) approach.
A better solution would be instead to make students work in groups, where students in the group use the same hardware.
This would reduce the amount of equipment needed and facilitate learning. 
Following this analogy, the class of students is a set of solutions proposed by the evolutionary optimizer, and the hardware is the environment.
The goal is to minimize the number of interactions with the environment in order to reduce the cost of the learning process.

Our experimental results on six RL tasks confirm that our approach can improve the efficiency of learning, ameliorating convergence while reducing, at the same time, the number of training episodes.
Moreover, in some cases, it can even improve the quality of the solutions with respect to the baseline.


The rest of the paper is structured as follows.
The next section describes the related work in the fields of social learning and IRL.
Then, we describe the proposed method in \Cref{sec:method}, and we present the experimental results in \Cref{sec:results}. Finally, in \Cref{sec:conclusions} we draw the conclusions of this work and list potential future research directions.


\section{Related work}
\label{sec:related_work}
In the following, we briefly summarize the literature on social learning and interpretable RL.

\subsection{Social Learning}
Social learning in animals and humans has been widely studied.
This paradigm of learning seems to be widespread, especially among primates.
In fact, the ``social brain'' hypothesis \cite{dunbar_social_1998} states that the size of primates' brains reflects the computational power needed for their complex social systems.
There is also evidence that social life improves the cognitive capabilities of individuals \cite{whiten_evolution_2007}.
Another hypothesis, called the ``costly information hypothesis'', states that animals use social learning when individual information is hard or costly to obtain \cite{zentall_social_1988,webster_social_2008}; furthermore, many studies observed that social animals prefer social learning over individual learning \cite{van2011social,Franz2009RapidEO}.
Interestingly, in many cases, it has been observed that using social learning leads to better performance, even when individual information is not costly or hard to acquire \cite{rendell_why_2010,stockwell_group_2017,kalkstein_social_2016}.

The idea of learning in groups is not novel in the field of Machine Learning (ML). Wang et al. \cite{junbo_wang_collaborative_2022} surveyed the state of the art in collaborative ML. One of the main paradigms in this area is currently that of Federated Learning (FL) \cite{li2020federated}. To some extent, our approach may look similar to FL, in that there are several agents that learn simultaneously and share some piece of information. However, unlike FL where multiple agents update the same model while each agent acts on its own data (e.g., due to privacy concerns), in our approach all the agents act on the \emph{same} instance of the environment during the collaborative phase.
Moreover, in our model agents do not learn by explicitly \emph{exchanging} knowledge (e.g., in the form of updated weights, as in FL). Instead, they learn simultaneously by \emph{acting} together in the same environment.

Furthermore, while our approach shares similarities with ensemble-based Q-Learning \cite{wang_adaptive_nodate,chen_ucb_2017,chen_randomized_2021}, it is fundamentally different.
In fact, while in ensemble Q-Learning all the agents have the same structure, in our case each agent has its own state-decomposition function, potentially leading to extremely different Q functions.
Moreover, another important difference between ensemble-based Q-Learning and our methodology is that while in the former the ``collective'' learning corresponds to the whole learning phase, in SIRL the collaborative phase is functional to reducing the training episodes required to train the agents while simultaneously enhancing their performance, but does not constitute the whole learning process.

Besides FL and ensemble-based Q-Learning, there are other works that apply social or collaborative approaches in ML.
\cite{Yaman_2022} study the effects of different meta-control strategies (i.e., strategies used to switch between various social learning modes) on the performance of agents in social games.
In \cite{zheng2020cooperative}, the authors propose a collaborative framework for deep RL. While our proposal shares some similarities with this approach, it is also significantly different.
In fact, in \cite{zheng2020cooperative} the knowledge is transferred between agents in a hierarchical fashion, while in SIRL the agents learn \emph{together} how to solve a task.
In \cite{team2022learning}, the authors study a domain in which a deep RL model can perform social learning from online human-generated data.
Finally, a growing trend in the field is to train large models in a collaborative fashion, to avoid the costs of a large, centralized infrastructure \cite{ryabinin_towards_2020,kijsipongse_hybrid_2018,medha_atre_distributed_2021,michael_diskin_distributed_2021,alexander_borzunov_training_2022}.

\subsection{Interpretable Reinforcement Learning}
The topic of trustworthiness in AI is an emerging area of research \cite{barredo_arrieta_explainable_2020}.
There are two main directions that try to address this issue, namely Explainable AI (XAI) and Interpretable AI (IAI). While the main focus of XAI is to produce \emph{a posteriori} explanations about opaque models, IAI is mostly about constructing models that are explainable \emph{by design}. In fact, even though the XAI field has been quickly growing in the past decade, some critics argued that XAI methods are not suitable for high-stakes scenarios \cite{rudin_stop_2019,rudin_interpretable_2021}.
However, the field of IAI (which some authors consider as a subfield of XAI) is somehow moving slower, which makes it harder for these methods to compete with non-interpretable state-of-the-art models.

Regarding IRL, there are mainly five lines of work in the field.
The first, oldest line of work makes use of DTs, trained with well-known algorithms for DT induction, modified in order to take into account the rewards received by the environment \cite{mccallum1996reinforcement,pyeatt_computer_nodate,saghezchi_multivariate_nodate,uther_tree_nodate,hwang_self_2006,hwang_induced_2012,hwang_tree-like_2007}.
While promising, these approaches suffer from the curse of dimensionality, making it hard to scale to larger RL problems.
Moreover, some of them are specifically tailored to specific types of state/action spaces, which limits their general applicability.

Another line of work consists in using small, easy-to-inspect neural networks \cite{liang2016state,malagon2019evolving}.
While these approaches are convenient, due to the fact that they can leverage most of the deep RL algorithms, their interpretability vanishes quickly as the state space or the action space grows.
In fact, several works that try to quantify interpretability \cite{virgolin_learning_2020,barcelo_model_2020} state that the number of operations performed by the model is a crucial indicator of the model interpretability.

Differentiable DTs (in short, DDTs) \cite{frosst_distilling_2017} are special DTs that do not use hard splits.
Instead, each split uses a sigmoid to weigh each of the two branches of the split.
Thus, the final decision of the DT is a weighted sum of its leaves.
In \cite{silva_optimization_2020}, the authors use DDTs for IRL.
The results show that the approach is able to achieve very good performance on various RL tasks including a real-world scenario. However, when discretizing DDTs into traditional DTs (in order to have interpretable solutions), the authors observe a significant reduction in performance.

While all the approaches mentioned above try to learn interpretable policies by making interpretable agents interact with the environment, another line of work aims to distill Non-Linear DTs (NLDTs) from deep neural networks \cite{dhebar_interpretable-ai_2020}.
Since this approach makes use of pre-trained deep models, it is very fast to train and allows customizing the properties of the induced DTs.
However, NLDTs make use of complex hyperplanes that make the interpretation of these solutions hard.

Finally, the last line of work combines population-based approaches, which optimize DT structures, and Q-Learning, which learns the Q-values for the leaves \cite{custode2023evolutionary,custode2021co}.
While this approach achieves performance that is comparable to the non-interpretable state of the art in several benchmarks, it requires a large number of interactions with the environment, reducing the competitiveness of this method w.r.t. the non-interpretable state of the art.


\section{Method}
\label{sec:method}
\begin{figure*}[ht!]
 \centering
 \resizebox{0.9\textwidth}{!}{
 \begin{tikzpicture}[>=latex,auto,node distance=1.5cm,main node/.style={circle,draw,font=\sffamily},]
 \node[rectangle, draw=cbgreen, rounded corners, thick, minimum width=3cm, minimum height=2cm] (opt) at (-4.3, 0) {\large \shortstack[c]{Evolutionary \\ optimizer}};
 \node[rectangle, draw=cbblue, fill=cbblue!10!white, rounded corners, thick, minimum width=7cm, minimum height=4.5cm, text depth=4.5cm] (coll) at (1.85, 0) {Collaborative phase};
 \node[] (t1) at (-0.5,1) {\usebox{\mytree}};
 \node[] (t2) at (-0.5,0) {\usebox{\mytree}};
 \node[] (t3) at (-0.5,-1) {\usebox{\mytree}};

 \node[draw, rectangle] (shared_env) at (4.3,0) {\includegraphics[width=1.5cm, height=2cm, trim=30 0 30 0, clip]{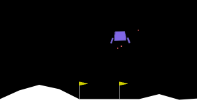}};
 \node[draw, rectangle, rounded corners] (aggregation) at (2,0) {\large Voting};

 \node[] () at (0.8,1) {$a_1$};
 \node[] () at (0.8,0.2) {$a_2$};
 \node[] () at (0.8,-1) {$a_3$};
 \draw[->] (opt.east) -- (coll.west);
 \draw[->] (t1.east) + (-0.1,-0.2) -| (aggregation.north);
 \draw[->] (t2.east) + (-0.1,0) -- (aggregation.west);
 \draw[->] (t3.east) + (-0.1,0.2) -| (aggregation.south);
 \draw[->] (aggregation.east) -- node[above, midway] {$a$} (shared_env.west);
 \draw[->] (-0.5,-1.92cm) -- (-1.5cm,-1.92cm) -- (-1.5cm, 0) |- (-1.25, 0);
 \draw[->] (shared_env.north) -- + (0,0.8cm) -- node[below] {$s, r, s'$} (-0.5,1.92cm) -- (t1.north);
 \draw[->] (shared_env.south) -- + (0,-0.8cm) -- node[above] {$s, r, s'$} (-0.5,-1.92cm) -- (t3.south);
 
 \node[rectangle, draw=cborange, fill=cborange!10!white, rounded corners, thick, minimum width=4.5cm, minimum height=4.5cm, text depth=4.5cm] (ind) at (8.9, 0) {Individual phase};
 \draw[->] (coll.east) -- (ind.west);
 \node[] (it1) at (7.5,1) {\usebox{\mytree}};
 \node[] (it2) at (7.5,0) {\usebox{\mytree}};
 \node[] (it3) at (7.5,-1) {\usebox{\mytree}};
 \node[draw, rectangle] (ienv1) at (10.3,1) {\includegraphics[width=0.7cm, height=0.7cm, trim=30 0 30 0, clip]{imgs/ll.png}};
 \node[draw, rectangle] (ienv2) at (10.3,0) {\includegraphics[width=0.7cm, height=0.7cm, trim=30 0 30 0, clip]{imgs/ll.png}};
 \node[draw, rectangle] (ienv3) at (10.3,-1) {\includegraphics[width=0.7cm, height=0.7cm, trim=30 0 30 0, clip]{imgs/ll.png}};
 \draw[->] ([yshift=0.1cm] ienv1.west) -- node[above, midway] {$s,r,s'$} ([yshift=0.1cm] it1.east);
 \draw[->] ([yshift=0.1cm] ienv2.west) -- node[above, midway] {$s,r,s'$} ([yshift=0.1cm] it2.east);
 \draw[->] ([yshift=0.1cm] ienv3.west) -- node[above, midway] {$s,r,s'$} ([yshift=0.1cm] it3.east);
 \draw[->] ([yshift=-0.1cm] it1.east) -- node[below, midway] {$a$} ([yshift=-0.1cm] ienv1.west);
 \draw[->] ([yshift=-0.1cm] it2.east) -- node[below, midway] {$a$} ([yshift=-0.1cm] ienv2.west);
 \draw[->] ([yshift=-0.1cm] it3.east) -- node[below, midway] {$a$} ([yshift=-0.1cm] ienv3.west);
 \draw[->] (ind.south) -- + (0, -0.5cm) -| (opt.south);
 \end{tikzpicture}
 }
 \caption{Graphical representation of the proposed SIRL approach.}
 \label{fig:mainfig}
\end{figure*}

The proposed method, which we call Social Interpretable RL (SIRL), is a general approach composed of two phases: a collaborative phase, and an individual phase.
During the collaborative phase, the agents learn \emph{together} how to solve the problem.
In the individual phase, instead, they interact \emph{separately} with the environment, with the goal of refining their policies.
Our approach, applied to IRL, is summarized graphically in \Cref{fig:mainfig}.

Basically, the rationale of our proposal is to exploit the possibility of off-policy algorithms to learn from transitions that are not strictly related to the policy of the agents. In order to do that, we extend the method from \cite{custode2023evolutionary} by implementing a social learning mechanism in the evaluation loop. 
The method from \cite{custode2023evolutionary} divides the IRL problem into two subproblems, namely:
\begin{enumerate}[leftmargin=*,label={(\arabic*)}]
 \item partitioning the state space ($\mathcal{S}$) into semantically similar groups of states;
 \item finding the optimal action for each partition defined in the previous step. 
\end{enumerate}
In \cite{custode2023evolutionary}, the problem is solved by using a two-level optimization approach. Subproblem (1) is solved by using Grammatical Evolution (GE) \cite{goos_grammatical_1998} in an outer loop to search for DTs that perform satisfactory partitions of $\mathcal{S}$, while subproblem (2) is solved by performing Q-Learning \cite{watkins1989learning} (i.e., in an inner loop) on the leaves of each DT.

We integrate the proposed SIRL into the algorithm proposed in \cite{custode2023evolutionary} as follows.
Subproblem (1) is solved as in the original method, i.e., using GE to search for DT structures.
Subproblem (2), instead, is solved by using the two phases of SIRL.
In the collaborative phase, all the agents from the population propose an action based on the current state.
Then, voting is performed on the proposed actions, and the resulting action is deployed to the environment.
Finally, all the agents observe the resulting transition $(s, a, r, s')$ and update their leaves, where: $s \in \mathcal{S}$ is the current state; $a \in \mathcal{A}$ is an action from the environment action space; $r = \mathcal{R}(s, a, s')$ is the reward given by the environment in response to the transition; and $s' \in \mathcal{S}$ is the next state.
In the following phase, i.e., the individual phase, each tree performs Q-learning on its own, and the average of its scores is used as fitness for that solution in the GE algorithm.
The loop is iterated for a number of iterations (i.e., generations) $g$, which is a hyperparameter.


\begin{algorithm}[ht!]
\caption{Training of DTs via Social Interpretable RL}
\label{alg:main}
\KwData{$p > 0, g > 0, e_c \geq 0, e_i > 0$}
\KwResult{$T^*$}
$T^* \gets NULL$ \;
\For{$gen \in [0, g]$}{
\tcp{Initialize or update pop}
\uIf{$gen = 0$}{
    $dts \gets initPop(p)$\tcp*{Initialize}
  }
  \Else{
     $dts \gets updatePop(dts, scores)$\tcp*{Update}
 }
 $collabPhase(dts, e_c)$\tcp*{\Cref{alg:collab}}
 $scores \gets []$\;
 \For{$dt \in dts$}{
 $scores \gets scores + [indPhase(dt, e_i)]$\;
 }
 $T \gets getBest(dts, scores)$\;
 \If{$isBetter(T, T^*)$} {
 $T^* \gets T$\tcp*{Update best}
 }
}
\Return $T^*$\;
\end{algorithm}

\begin{algorithm}[ht!]
\caption{Collaborative Phase}
\label{alg:collab}
\SetNoFillComment
\KwData{$dts, e_c \geq 0$}
\tcc{Iterate over episodes}
\For{$episode \in [1, e_c]$}{
 $s \gets getState()$\tcp*{Retrieve state}
 $done \gets false$\;
 \tcc{Iterate over the episode}
 \While{$!done$}{
 $actions \gets []$\;
 \For{$dt \in dts$}{
 \tcp{Collect agents' actions}
 $actions \gets actions + [dt.getAction(s)]$\;
 }
 $a \gets randomSel(actions)$\tcp*{Voting}
 $s', r, done \gets executeAction(a)$\;
 \For{$dt \in dts$}{
 $qlearn(dt, s, a, r, s')$\tcp*{Q-Learning}
 }
 }
}
\end{algorithm}

\begin{algorithm}[ht!]
\caption{Individual Phase}
\label{alg:individual}
\SetNoFillComment
\KwData{$dts, e_c \geq 0$}
\tcc{Iterate over individuals}
\For{$dt \in dts$}{
\tcc{Iterate over episodes}
\For{$episode \in [1, e_c]$}{
 $s \gets getState()$\tcp*{Retrieve state}
 $done \gets false$\;
 \tcc{Iterate over the episode}
 \While{$!done$}{
 $a \gets dt.getAction(s)$\;
 }
 $s', r, done \gets executeAction(a)$\;
 $qlearn(dt, s, a, r, s')$\tcp*{Q-Learning}
 }
}
\end{algorithm}

The pseudocode of the algorithm is shown in \Cref{alg:main}, where $p$ is the size of the population for GE, $g$ is the number of iterations of the optimization process, $e_c$ is the number of collaborative episodes, $e_i$ is the number of individual episodes, and $T^*$ is the best tree found during the process.
\Cref{alg:collab}, instead, shows the functioning of the collaborative phase, which takes in input a list of DTs ($dts$), and the number of collaborative episodes $e_c$.
Finally, \Cref{alg:individual} contains the pseudocode for the individual phase of our approach.

As shown in \Cref{alg:collab}, the voting phase randomly chooses one of the actions proposed by the agents.
This is due to the fact that this mechanism should automatically balance exploration and exploitation.
In fact, in the initial phase, the distribution of the actions proposed by the agents is close to a uniform distribution.
Then, while the training progresses, we expect the emergence of \emph{consensus} between the agents.
This means that, at this point, the choice will be heavily biased toward the most widely chosen action, while still allowing to explore by choosing actions that were selected by a minority of the agents.
Finally, as training progresses and agents learn the optimal state-action mapping, we expect the probability of choosing the most chosen action to reach a plateau.
This is due to the fact that, as training progresses, individuals with a good tree structure will learn a (sub-)optimal state-action mapping, which increases the probability of choosing the optimal action, while individuals with a bad tree structure will still behave randomly. This tells us that the probability of choosing the most ``common'' action can only increase during training.
In preliminary experiments, we tested alternative voting mechanisms such as ``follow the leader'' (i.e., choose the action of the ``best'' agent from a small initial individual phase) and majority voting, but we could not observe any significant differences in terms of scores w.r.t. random voting.

Note that, in our method, agents do not share their experiences. In fact, in order to share experiences, each agent would need to explore the environment on its own. On the other hand, in our collaborative phase, the exploration is conducted by all the agents at the same time. This is due to the fact that, after each action (which has been decided collectively), all the agents are updated with the very same transition. Thus, in our case, the exploration is guided by all the agents at the same time. This, given a fixed computational cost, gives more control to each agent on the transitions that it experiences (w.r.t. sharing experiences).
Moreover, our approach can be seen as a mix of online and offline learning.
In fact, while the baseline method \cite{custode2023evolutionary} solely performs online learning, other methods (e.g., \cite{dhebar_interpretable-ai_2020}) only perform offline learning and experience sharing performs a mix of online and offline learning (with hard-coded proportions); our method allows exploring the continuum between online and offline RL without requiring hard-coded proportions between the amount of data learned online and that learned offline.
In fact, in our approach, each agent has a direct influence on the sampling probability for each of the actions, indirectly influencing the action chosen for each step.

From the point of view of computational complexity, our approach is very convenient especially when the optimizer has to evaluate a large number of agents at each iteration.
In fact, having a set of $p$ agents, $g$ iterations, and $e$ episodes, the computational cost of the process from \cite{custode2023evolutionary} would be $O(p\cdot g\cdot e)$, where each agent experiences $e$ episodes.
Instead, with SIRL, using $e_c$ collaborative episodes and $e_i$ individual episodes, we obtain a computational complexity of $O(p\cdot g\cdot e_i + g\cdot e_c)$.
Hence, in SIRL each agent experiences $e_c + e_i$ episodes.

The advantages introduced by SIRL are twofold.
Firstly, we can reduce the computational cost of the training by balancing $e_i$ and $e_c$ in such a way that:
\begin{equation}
 e_i + \frac{e_c}{p} < e.
\end{equation}
Secondly, we can improve the quality of the agents produced by carefully tuning $e_c$ and $e_i$ in such a way that the number of episodes seen by each agent is higher than those seen without using SIRL, i.e.:
\begin{equation}
 e_i + e_c > e.
\end{equation}
Of course, it is important to note that, with our approach, it is possible to have \emph{simultaneously} both advantages, by carefully tuning $e_c$ and $e_i$.
For instance, one may want to maximize the number of episodes experienced in the collective phase (i.e., maximizing $e_c$), while ensuring that the number of episodes seen during the individual phase is enough to have a good estimate of the average performance of the individual (e.g., using $e_i=3$ for environments with a relatively small amount of noise, while using larger values for more noisy environments).

Note that our approach assumes that the main bottleneck of the training process is the simulation of the environment (i.e., the computation of the ``steps'' in the environment). If this assumption does not hold for a given environment, then this approach is not going to be more efficient than the baseline method. It is important to note, however, that the inference time of interpretable DTs easily becomes negligible, due to the fact that the tree depth is limited in order to achieve interpretability and that inference time for DTs is logarithmic w.r.t. the number of nodes.

Finally, it is important to note that both the collaborative phase and the individual phase are parallelized.
For the individual phase, the parallelization is straightforward: each DT is evaluated in a separate job.
On the other hand, for the collaborative phase, we port a widely-known parallelization scheme from neural-network-based RL approaches \cite{jaderberg2019human}, by making the agent experience several episodes in parallel and averaging the leaves at the end of the collaborative phase (see the Supplementary Material).


\section{Results}
\label{sec:results}
This section presents the experimental setup and the results obtained using SIRL.
All the experiments were executed on an HPC, where each run (i.e., each optimization process on a given task) was allocated $30$ cores and 4GB of RAM. Our code, which we make publicly available on DagsHub\footnote{\url{https://dagshub.com/leocus4/sociallearningdts}} together with the experimental results, is based on the DEAP library \cite{fortin_deap_2012}, and uses the environments implemented in the Gymnasium library \cite{brockman2016openai}.

For each setting (i.e., each method on a given task), we perform $10$ independent runs, in order to assess the statistical repeatability of the results.

We test our method on six widely-used, well-known benchmarks from the Gymnasium suite: InvertedPendulum, Swimmer, Reacher, Hopper, and Walker2d from the MuJoCo suite, and LunarLander from the Box2D suite.

\begin{figure*}[ht!]
 \centering
 \includegraphics[width=0.99\textwidth]{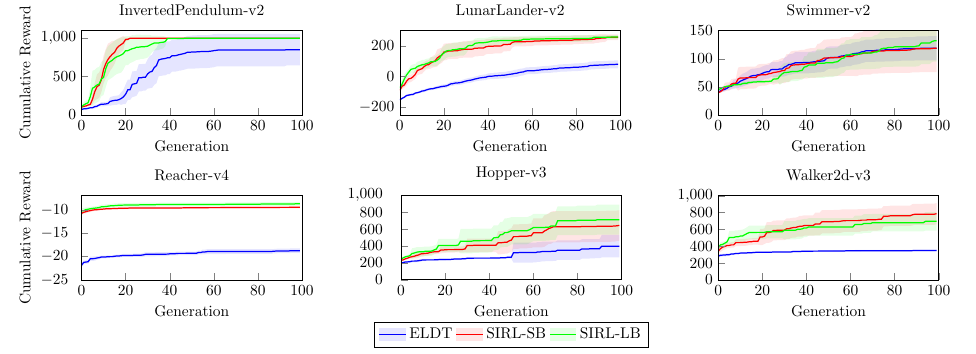}
 \caption{Scores (the higher, the better) obtained by the best agent, at each iteration of the outer loop (i.e., a generation), for the population-based training of interpretable agents. The solid line represents the mean value, while the shaded area represents the $95\%$ CI.
 }
 \label{fig:dts_trends}
\end{figure*}

\Cref{fig:dts_trends} shows the mean score obtained by the best agent at each iteration of the optimization process (see Supplementary Material for the hyperparameters).
In this figure, we compare the original approach \cite{custode2023evolutionary} (ELDT) with two configurations of SIRL: (1) SIRL-SB (standard budget), i.e., SIRL using the same number of interactions with the environment (except for LunarLander-v2, where ELDT needed a much greater number of episodes to obtain reasonable performance); and (2) SIRL-LB (low budget): i.e., SIRL using a substantially smaller number of interactions with the environment (from $43$\% to $76$\% less episodes, depending on the task, see Supplementary Material).
We observe that, in both InvertedPendulum and LunarLander, SIRL has a clear advantage over the original ELDT method.
In fact, the agents trained with SIRL achieve better scores and converge faster.
Interestingly, there is no significant difference between SIRL-SB and SIRL-LB, indicating that the number of episodes seen by each agent in the LB case may already be enough to achieve satisfactory performance.
On the other hand, in the Swimmer task, SIRL does not bring any additional advantage to ELDT, besides reducing the computational cost.
This may indicate that, in this task, the optimal structure of the DT is harder to find, and thus more generations (i.e., iterations of the outer loop) are needed to find good architectures. This shows that, while SIRL can improve the performance of the inner loop, it does not affect significantly the outcome of the outer loop.

\begin{figure*}[ht!]
 \centering
 \includegraphics[width=0.99\textwidth]{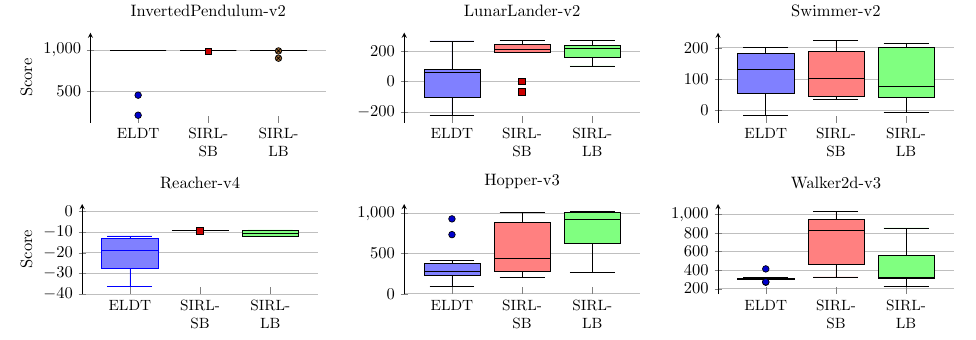}
 \caption{Scores (the higher, the better) obtained by the best agents found by each tested method in $10$ runs, tested on $100$ unseen episodes.}
 \label{fig:dts_boxplot}
 \vspace{-1cm}
\end{figure*}

In \Cref{fig:dts_boxplot}, instead, we show the boxplots of the scores obtained by the best agents produced in each of the $10$ runs, when testing them on $100$ unseen episodes.
We observe that SIRL always achieves comparable or better results than the baseline method (ELDT).
Surprisingly, we observe that SIRL-LB has often a lower variance than SIRL-SB, which may indicate that one may have to find a balance between $e_c$ and $e_i$ to maximize performance.

To confirm the statistical significance of the results, we perform a one-way ANOVA test (with the Tukey post-hoc procedure) on the results shown in \Cref{fig:dts_boxplot}.
We use a confidence level $\alpha = 0.05$.
The results of the statistical tests are shown in the Supplementary Material.
From this analysis, it emerges that the null hypothesis can be rejected in three cases over six.
This confirms that, in the worst case, SIRL achieves the same scores while using significantly fewer episodes, while in the best case, it simultaneously reduces the number of episodes and improves performance.

\begin{figure}[ht!]
 \centering
 \includegraphics[width=0.9\columnwidth]{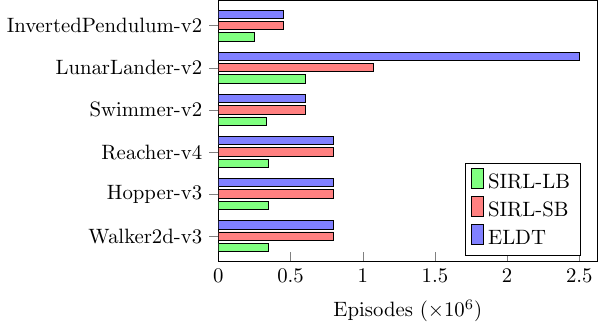}
 \caption{Number of episodes used by each of the algorithms under comparison.}
 \label{fig:budget}
\end{figure}

In \Cref{fig:budget}, we show the number of episodes used by each of the three approaches, which demonstrates the computational advantage of SIRL.

\newcommand{\bestint}[1]{\underline{#1}}
\newcommand{\bestall}[1]{\underline{\underline{#1}}}
\begin{table*}[ht!]
 \renewcommand{\arraystretch}{1.1}
 \centering
 \caption{
 Comparison of the proposed SIRL approaches with SOTA opaque/interpretable methods with respect to: mean $\pm$ std. dev. score across $100$ episodes (the higher the better); number of episodes for training (the lower the better), reported as a proxy for the training cost; and complexity, expressed in terms of number of MAC operations per episode (the lower the better), reported as a proxy for interpretability \cite{barcelo_model_2020}. The standard deviation for the score is shown only where available in the original papers. The scores for the SIRL approaches refer to the best agents (in terms of score), among the agents found in 10 independent runs, respectively for SIRL-SB and SIRL-LB. ``$\ddagger$'' indicates a lower bound estimate for the number of episodes.
 The baseline (ELDT) is the method presented in \cite{custode2023evolutionary}. However, the results have been recomputed in order to have a comparable budget w.r.t. SIRL-SB.
 Single (double) underlining indicates the best result among interpretable (all) methods. Asterisks denote a statistically significant difference in score between the SIRL approaches and the baseline (Welch T-Test, $\alpha=0.05$), namely: ``$^{**}$'': $p \leq 0.01$; ``$^{***}$'': $p \leq 0.001$; ``$^{****}$'': $p \leq 0.0001$. ``$^{\mathrm{ns}}$'' denotes a non-significant difference.
 } 
 \label{tab:sota}
 \resizebox{0.8\textwidth}{!}{
 \let\mathpm=\pm
 \let\pm=&
 \begin{tabular}{l l l r @{ $\mathpm$ } r @{\hskip 0.3in} c @{\hskip 0.3in} c c c} \toprule
 \textbf{Environment} & \textbf{Method} & \textbf{Type} & \multicolumn{2}{c}{\textbf{Score}} & & & \textbf{Episodes} & \textbf{Complexity} \\
 \midrule
 \multirow{7}[1]{*}{InvertedPendulum-v2}
 & TD3 \cite{patel2022temporally} & Opaque & ${1000.00}$ \pm ${0.00}$ & & & \bestall{$1.0\cdot 10^3$}$\ddagger$ & $1.2 \cdot 10^8$ \\
 & TempoRL \cite{patel2022temporally} & Opaque & ${1000.00}$ \pm ${0.00}$ & & & \bestall{$1.0\cdot 10^3$}$\ddagger$ & $1.3 \cdot 10^8$ \\
 & TLA-O \cite{patel2022temporally} & Opaque & ${1000.00}$ \pm ${0.00}$ & & & \bestall{$1.0\cdot 10^3$}$\ddagger$ & $2.6 \cdot 10^7$ \\
 & TLA-C \cite{patel2022temporally} & Opaque & ${1000.00}$ \pm ${0.00}$ & & & \bestall{$1.0\cdot 10^3$}$\ddagger$ & $2.6 \cdot 10^7$ \\ 
 & ELDT (baseline) & Interpretable & ${1000.00}$ \pm ${0.00}$ & \multirow{2}{*}{\hspace{-1.5em}$\left.\begin{array}{l} \\ \\ \end{array}\right]^{\mathrm{ns}}$} & \multirow{3}{*}{\hspace{-2em}$\left.\begin{array}{l} \\ \\ \\ \end{array}\right]^{\mathrm{ns}}$} & $4.5 \cdot 10^5$$\phantom{\ddagger}$ & $2.0\cdot 10^3$ \\
 \cmidrule(l{0.0in}r{0.3in}){2-5} \cmidrule(l{0.2in}r{0.2in}){7-9}
 & SIRL-SB (ours) & Interpretable & ${1000.00}$ \pm ${0.00}$ & & & ${4.5 \cdot 10^5}$$\phantom{\ddagger}$ & $5.0\cdot 10^3$ \\
 & SIRL-LB (ours) & Interpretable & ${1000.00}$ \pm ${0.00}$ & & & \bestint{$2.5 \cdot 10^5$}$\phantom{\ddagger}$ & \bestall{$1.0 \cdot 10^3$}\\
 \midrule
 \multirow{9}[1]{*}{LunarLander-v2}
 & Value difference \cite{xu_deep_2020} & Opaque & ${248.20}$ \pm ${21.00}$ & & & ${1.0\cdot 10^4}$$\phantom{\ddagger}$ & $1.1\cdot 10^7$ \\
 & Advantage weighting \cite{peng_advantage-weighted_2019} & Opaque & ${229.00}$ \pm ${}$ & & & ${1.0\cdot 10^4}$$\ddagger$ & $9.7\cdot 10^6$ \\
 & Shallow neural network \cite{malagon2019evolving} & Interpretable & ${258.80}$ \pm ${}$ & & & ${7.0\cdot 10^5}$$\phantom{\ddagger}$ & $3.2\cdot 10^4$ \\
 & Differentiable DTs \cite{silva_optimization_2020} & Interpretable & ${-78.40}$ \pm ${32.20}$ & & & ~${1.0\cdot 10^4}$$\phantom{(\ddagger}$ & N/A\\
 & Non-Linear DTs \cite{dhebar_interpretable-ai_2020} & Interpretable & ${234.98}$ \pm ${22.25}$ & & & N/A & N/A\\
 & PW-Net \cite{kenny2023towards} & Interpretable & ${216.94}$ \pm ${16.92}$ & & & ~\bestall{$1.5\cdot 10^3$}$\phantom{(\ddagger}$ & ${3.0}\cdot 10^6$ \\
 & ELDT (baseline) & Interpretable & ${244.20}$ \pm ${82.53}$ & \multirow{2}{*}{\hspace{-1.5em}$\left.\begin{array}{l} \\ \\ \end{array}\right]^{***}$} & \multirow{3}{*}{\hspace{-2em}$\left.\begin{array}{l} \\ \\ \\ \end{array}\right]^{**}$} & $2.5\cdot 10^6\phantom{\ddagger}$ & $1.3\cdot 10^4$ \\
 \cmidrule(l{0.0in}r{0.3in}){2-5} \cmidrule(l{0.2in}r{0.2in}){7-9}
 & SIRL-SB (ours) & Interpretable & \bestall{$272.21$} \pm ${27.46}$ & & & $1.1\cdot 10^6\phantom{\ddagger}$ & $1.1\cdot 10^4$ \\
 & SIRL-LB (ours) & Interpretable & ${266.03}$ \pm ${41.03}$ & & & $6.0\cdot 10^5\phantom{\ddagger}$ & \bestall{$8.0\cdot 10^3$} \\
 \midrule
 \multirow{7}[1]{*}{Swimmer-v2}
 & CEM-RL \cite{pourchot2019cemrl,zheng2020cooperative} & Opaque & \bestall{$274.00$} \pm ${118.00}$ & & & \bestall{$1.0\cdot 10^3$}$\ddagger$ & $1.25 \cdot 10^8$ \\
 & CSPC \cite{zheng2020cooperative} & Opaque & ${261.00}$ \pm ${117.00}$ & & & \bestall{$1.0\cdot 10^3$}$\ddagger$ & N/A\\
 & PPO \cite{zheng2020cooperative} & Opaque & ${68.00}$ \pm ${}$ & & & \bestall{$1.0\cdot 10^3$}$\ddagger$ & $4.9\cdot 10^6$ \\
 & SAC \cite{zheng2020cooperative} & Opaque & ${44.00}$ \pm ${}$ & & & \bestall{$1.0\cdot 10^3$}$\ddagger$ & $6.9 \cdot 10^7$ \\
 & ELDT (baseline) & Interpretable & ${202.62}$ \pm ${2.61}$ & \multirow{2}{*}{\hspace{-1.5em}$\left.\begin{array}{l} \\ \\ \end{array}\right]^{****}$} & \multirow{3}{*}{\hspace{-2em}$\left.\begin{array}{l} \\ \\ \\ \end{array}\right]^{****}$} & $6.0\cdot 10^5\phantom{\ddagger}$ & $7.0\cdot 10^3$ \\
 \cmidrule(l{0.0in}r{0.3in}){2-5} \cmidrule(l{0.2in}r{0.2in}){7-9}
 & SIRL-SB (ours) & Interpretable & $\bestint{223.24}$ \pm ${5.09}$ & & & $6.0\cdot 10^5\phantom{\ddagger}$ & $1.6\cdot 10^4$ \\
 & SIRL-LB (ours) & Interpretable & ${215.44}$ \pm ${4.00}$ & & & \bestint{$3.3\cdot 10^5$}$\phantom{\ddagger}$ & \bestall{$5.0\cdot 10^3$} \\
 \midrule
 \multirow{7}[1]{*}{Reacher-v4}
 & POP3D \cite{chu2018policy} & Opaque & ${-4.29}$ \pm ${}$ & & & ${1.0\cdot 10^4}$$\ddagger$ & $1.5\cdot 10^6$ \\
 & PPO \cite{chu2018policy} & Opaque & ${-5.00}$ \pm ${}$ & & & ${1.0\cdot 10^4}$$\ddagger$ & $5.1\cdot 10^6$ \\
 & DDPG \cite{wu2022td3} & Opaque & \bestall{$-4.09$} \pm ${}$ & & & \bestall{$1.0\cdot 10^3$}$\ddagger$ & $6.9\cdot 10^7$ \\
 & A-TD3 \cite{wu2022td3} & Opaque & ${-4.77}$ \pm ${}$ & & & \bestall{$1.0\cdot 10^3$}$\ddagger$ & $6.9\cdot 10^7$ \\
 & ELDT (baseline) & Interpretable & ${-11.96}$ \pm ${5.05}$ & \multirow{2}{*}{\hspace{-1.5em}$\left.\begin{array}{l} \\ \\ \end{array}\right]^{****}$} & \multirow{3}{*}{\hspace{-2em}$\left.\begin{array}{l} \\ \\ \\ \end{array}\right]^{****}$} & $8.0\cdot 10^5\phantom{\ddagger}$ & $5.0\cdot 10^3$ \\
 \cmidrule(l{0.0in}r{0.3in}){2-5} \cmidrule(l{0.2in}r{0.2in}){7-9}
 & SIRL-SB (ours) & Interpretable & ${-9.10}$ \pm ${2.78}$ & & & $8.0\cdot 10^5\phantom{\ddagger}$ & \bestall{$2.0\cdot 10^3$} \\
 & SIRL-LB (ours) & Interpretable & $\bestint{-8.71}$ \pm ${2.44}$ & & & \bestint{$3.5\cdot 10^5$}$\phantom{\ddagger}$ & \bestall{$2.0\cdot 10^3$} \\
 \midrule
 \multirow{8}[1]{*}{Hopper-v3}
 & TD3 \cite{wu2022td3} & Opaque & ${3390.79}$ \pm ${}$ & & & \bestall{$1.0\cdot 10^3$}$\ddagger$ & $7.0\cdot 10^7$ \\
 & A-TD3 \cite{wu2022td3} & Opaque & ${3307.78}$ \pm ${}$ & & & \bestall{$1.0\cdot 10^3$}$\ddagger$ & $7.0\cdot 10^7$ \\
 & MDC-SAN \cite{zhang2022multi} & Opaque & \bestall{$3636.00$} \pm ${}$ & & & \bestall{$1.0\cdot 10^3$}$\ddagger$ & $7.0\cdot 10^7$ \\
 & Tree GP \cite{videau2022multi} & Interpretable & ${999.19}$ \pm ${}$ & & & $3.0\cdot 10^6\phantom{\ddagger}$ & N/A\\
 & Linear GP \cite{videau2022multi} & Interpretable & ${949.27}$ \pm ${}$ & & & $3.0\cdot 10^6\phantom{\ddagger}$ & N/A\\
 & ELDT (baseline) & Interpretable & ${939.55}$ \pm ${196.52}$ & \multirow{2}{*}{\hspace{-1.5em}$\left.\begin{array}{l} \\ \\ \end{array}\right]^{****}$} & \multirow{3}{*}{\hspace{-2em}$\left.\begin{array}{l} \\ \\ \\ \end{array}\right]^{****}$} & $8.0\cdot 10^5\phantom{\ddagger}$ & $1.2\cdot 10^4$ \\
 \cmidrule(l{0.0in}r{0.3in}){2-5} \cmidrule(l{0.2in}r{0.2in}){7-9}
 & SIRL-SB (ours) & Interpretable & ${1017.96}$ \pm ${0.64}$ & & & $8.0\cdot 10^5\phantom{\ddagger}$ & \bestall{$4.0\cdot 10^3$} \\
 & SIRL-LB (ours) & Interpretable & \bestint{$1026.28$} \pm ${0.62}$ & & & \bestint{$3.5\cdot 10^5$}$\phantom{\ddagger}$ & $2.3\cdot 10^4$ \\
 \midrule
 \multirow{6}[1]{*}{Walker2d-v3}
 & TD3 \cite{wu2022td3} & Opaque & ${4413.89}$ \pm ${}$ & & & \bestall{$1.0\cdot 10^3$}$\ddagger$ & $7.2\cdot 10^7$ \\
 & A-TD3 \cite{wu2022td3} & Opaque & ${4136.88}$ \pm ${}$ & & & \bestall{$1.0\cdot 10^3$}$\ddagger$ & $7.2\cdot 10^7$ \\
 & MDC-SAN \cite{zhang2022multi} & Opaque & \bestall{$5566.00$} \pm ${}$ & & & \bestall{$1.0\cdot 10^3$}$\ddagger$ & $7.2\cdot 10^7$ \\
 & ELDT (baseline) & Interpretable & ${318.79}$ \pm ${28.04}$ & \multirow{2}{*}{\hspace{-1.5em}$\left.\begin{array}{l} \\ \\ \end{array}\right]^{****}$} & \multirow{3}{*}{\hspace{-2em}$\left.\begin{array}{l} \\ \\ \\ \end{array}\right]^{****}$} & $8.0\cdot 10^5\phantom{\ddagger}$ & \bestall{$2.0\cdot 10^3$} \\
 \cmidrule(l{0.0in}r{0.3in}){2-5} \cmidrule(l{0.2in}r{0.2in}){7-9}
 & SIRL-SB (ours) & Interpretable & \bestint{$982.38$} \pm ${73.71}$ & & & $8.0\cdot 10^5\phantom{\ddagger}$ & $4.0\cdot 10^3$ \\
 & SIRL-LB (ours) & Interpretable & ${722.73}$ \pm ${313.26}$ & & & \bestint{$3.5\cdot 10^5$}$\phantom{\ddagger}$ & $4.0\cdot 10^3$ \\
 \bottomrule
 \end{tabular}
 }
\end{table*}

Finally, we compare our results to the state of the art in \Cref{tab:sota}.
In three cases over six, we observe that SIRL obtains comparable or better performance w.r.t. the non-interpretable state of the art (InvertedPendulum, LunarLander, and Swimmer).
In the remaining three cases, while the performance obtained by SIRL cannot be compared to those of opaque RL methodologies, it significantly surpasses that of all the other interpretable RL methodologies.
This confirms that, with SIRL, the interpretable RL field is a step closer to matching non-interpretable RL methodologies.

The best DTs found by SIRL for each environment can be found in the Supplementary Material, along with their interpretation, which practically shows the advantage of using interpretable models.


\section{Conclusions}
\label{sec:conclusions}

Interpretability is a growing concern in the field of ML.
In fact, it is a key enabler for trustworthiness, as it allows a thorough understanding of a given model.
While some approaches to IRL have been proposed in recent years, one of the main issues with these approaches is their computational cost.
In this paper, we proposed a method that significantly reduces this cost, showing comparable, or even better performance by reducing the number of interactions with the environment from $43\%$ to $76\%$, depending on the task at hand.
In fact, the experimental results show that SIRL converges faster while finding solutions that have comparable or even better performance than the original method from \cite{custode2023evolutionary} and, in general, than other interpretable RL methodologies.

Future work will focus on: (1) scaling this method to larger populations and longer optimization processes; (2) studying the effect of the $(e_c,e_i)$ pairs on the performance, to understand the relationship between the two parameters; (3) designing more sophisticated social learning schemes that, in a single generation, interleave collaborative and individual phases in an iterative fashion, allowing for a continuous improvement of the voting strategy (e.g., by weighting the votes); (4) devising novel voting mechanisms for SIRL; (5) porting the proposed social learning approach to deep RL for hyperparameter optimization and Neural Architecture Search (see the preliminary results reported in the Supplementary Material); (6) using more sophisticated leaves (e.g., linear and non-linear models) for allowing better scaling w.r.t. the size of the action space; and (7) extending SIRL with novel mechanisms in order to improve sample efficiency w.r.t. the opaque state of the art.

\textbf{Limitations.} While the proposed approach showed promising performance, it suffers from limitations.
Firstly, this method currently works only with environments whose state space is composed of semantically ``meaningful'' variables.
By meaningful, we mean that each of the inputs has its own meaning and can be used to make decisions.
For instance, a state space composed of images does not contain ``meaningful'' variables, in that a single pixel may not contain enough information to make a decision.
This means that the models trained with our approach can use knowledge from the state space, but they cannot \emph{build} it (e.g., by extracting features), as done by neural networks used in deep RL.

Another limitation concerns the type and size of the action space.
In fact, when the action space becomes very large, the number of episodes required to thoroughly explore the value of each action grows significantly.
In fact, while SIRL is able to significantly reduce the number of episodes, the cost of training IRL models with our approach in such environments may still be prohibitive.
Finally, it is worth noting that, while IRL methods can be competitive with established deep RL algorithms, sometimes their performance is not comparable to that of state-of-the-art non-interpretable RL methods, as shown in the experiments on some of the MuJoCo environments.
However, the results from the literature and the present work should encourage research in this direction.


\begin{credits}
\subsubsection{\ackname} 
Funded by the European Union (project no. 101071179). Views and opinions expressed are however those of the author(s) only and do not necessarily reflect those of the European Union or EISMEA. Neither the European Union nor the granting authority can be held responsible for them.
\end{credits}


\clearpage
\bibliographystyle{splncs}
\bibliography{main.bib}


\appendix
\clearpage
\setcounter{section}{0}
\begin{center}
  {\Large \bfseries\boldmath
  Supplementary Material \par}
\end{center}

\section{Hyperparameters}

SIRL makes use of a two-level optimization approach, as presented in \cite{custode2023evolutionary}.
In the outer level, we use Grammatical Evolution \cite{goos_grammatical_1998} which, using the grammar shown in \Cref{tab:grammar}, translates a list of $l_g$ integer values (i.e., the genotype) into a Decision Tree (DT), i.e., the phenotype.
The translation process is shown in \Cref{alg:translation}.
The inner level, instead, uses Q-learning \cite{watkins1989learning} to learn the Q-values for each of the possible actions in the leaves of the DT.

Note that, for the continuous environments, a discretization of the action space must be performed.
The action space is discretized in the following way.
For each action, we use $7$ bins.
This means that, given an environment with $n_a$ continuous outputs, we use $7 \cdot n_a$ discrete actions.
When the DT chooses a discrete action, we convert it into a continuous action by initializing a vector $\mathbf{a} = \mathbf{0}$ and then selecting the chosen output $a_c$ based on the discrete action chosen by the DT, $a_{DT}$:
\begin{equation}
 a_c = \lfloor \frac{a_{DT}}{n_a}\rfloor
\end{equation}
Once we know what action has been selected, we can translate $a_{DT}$ into a continuous action as follows:
\begin{equation}
 a[a_c] = 2 \cdot \frac{a_{DT}~mod~7}{6} - 1
\end{equation}
The choice underlying the use of $7$ bins lies in the results shown in \cite{tang_discretizing_2020}.

The hyperparameters for all the methods used in the experimental phase are shown in \Cref{tab:hparams}.
All the values have been set as in \cite{custode2023evolutionary}, where they have been manually tuned for the tasks at hand and the computational budget available.
Note that the mutation probability refers to the probability of mutating a single individual, while the mutation rate controls how many codons from a single individual are mutated.

\begin{algorithm}[!ht]
\caption{Translation of a genotype into a phenotype.}
\label{alg:translation}
\SetNoFillComment
\KwData{$x$: a genotype}
\KwData{$grammar$: a dictionary}
\KwResult{$T$: a DT}
\tcc{Initialize the tree as a string with the starting symbol}
$tree \gets ``\langle start\rangle"$\;
\For{$n \in x$}{
 \If{$hasNonExpandedRules(tree)$}{
 \tcc{Retrieve the first non-expanded rule from the string}
 $nonExpandedRule \gets getFirstNonExpandedRule(tree)$\;
 \tcc{Retrieve all options from the grammar for the current rule}
 $options \gets grammar[nonExpandedRule]$\;
 \tcc{Choose one of the options according to the current integer. Use the modulo operator to ensure that the index is valid}
 $choice \gets options[n~mod~len(options)]$\;
 \tcc{Replace the first occurrence of the rule with the choice.}
 $tree \gets replaceFirst(nonExpandedRule, choice)$
 }
 \Else{
 $break$\;
 }
}
\Return $tree$\;
\end{algorithm}

\begin{table}[!ht]
 \centering
 \caption{Grammar used to translate genotypes into phenotypes. The $\mid$ symbol separates distinct options for the production of a rule. Rules are enclosed by angular brackets.}
 \label{tab:grammar}
 \begin{tabular}{l l} \toprule
 \textbf{Rule} & \textbf{Options} \\ \midrule
 start & $\langle if\rangle$ \\
 if & $Node(\langle condition\rangle, \langle action\rangle, \langle action\rangle)$ \\
 condition & $\langle const\rangle \cdot x_1 + \dots + \langle const\rangle \cdot x_N < \langle const\rangle$ \\
 action & Leaf $\mid \langle if\rangle$ \\
 const & $0 \mid \langle nonzero\rangle$\\
 nonzero & $[-10, 10]$ with step $0.1$\\
 \bottomrule
 \end{tabular}
\end{table}

\begin{table*}[!ht]
 \centering
 \caption{Hyperparameters used in the experimental phase. $\mathcal{U}$ denotes the uniform probability distribution.}
 \label{tab:hparams}
 \begin{tabular}{l l c} \toprule
 \textbf{Hyperparameter} & \textbf{Environment} & \textbf{Value} \\ \midrule
 \multirow{6}{*}{Individual episodes} & InvertedPendulum-v2 & 3 \\
 & LunarLander-v2 & 10 \\
 & Swimmer-v2 & 5 \\
 & Reacher-v4 & 5 \\
 & Hopper-v3 & 5 \\
 & Walker2d-v3 & 5 \\
 \hline
 \multirow{3}{*}{\shortstack{Collaborative parallel \\ episodes $e_p$}} & InvertedPendulum-v2 & 10 \\
 & LunarLander-v2 & 10 \\
 & Swimmer-v2 & 10 \\
 & Reacher-v4 & 10 \\
 & Hopper-v3 & 10 \\
 & Walker2d-v3 & 10 \\
 \hline
 \multirow{3}{*}{\shortstack{Collaborative iterations \\ $i$}} & InvertedPendulum-v2 & 100 \\
 & LunarLander-v2 & 100 \\
 & Swimmer-v2 & 80 \\
 & Reacher-v4 & 100 \\
 & Hopper-v3 & 100 \\
 & Walker2d-v3 & 100 \\
 \hline
 Q-table initialization & & \multicolumn{1}{c}{$\sim\mathcal{U}(-1, 1)$} \\
 Discount factor $\gamma$ & & \multicolumn{1}{c}{0.9} \\
 Exploration probability $\varepsilon$ & & \multicolumn{1}{c}{0.05} \\
 Learning rate $\alpha$ & & \multicolumn{1}{c}{0.1} \\
 Cross-over probability & & \multicolumn{1}{c}{0.1} \\
 Crossover operator & & \multicolumn{1}{c}{One point crossover} \\
 Mutation probability & & \multicolumn{1}{c}{0.9} \\
 Mutation operator & & \multicolumn{1}{c}{$\sim\mathcal{U}(0, 4e4)$}\\
 Mutation rate & & \multicolumn{1}{c}{0.05}\\
 Selection operator & & \multicolumn{1}{c}{Tournament (size 2)} \\
 Genotype length $l_g$ & & \multicolumn{1}{c}{1000} \\
 Grammar type & & \multicolumn{1}{c}{oblique} \\
 Population size & & \multicolumn{1}{c}{500} \\
 Number of generations & & \multicolumn{1}{c}{100} \\
 \bottomrule
 \end{tabular}
\end{table*}

\section{Computational environment}
The HPC used for our experimentation has the following specifications:
\begin{itemize}[leftmargin=*]
 \item CPUs: $142$ CPU calculation nodes for a total of $7.674$ cores
 \item Ram: $65$TB
 \item Theoretical peak performance (CPU): $422,7$ TFLOPs
\end{itemize}

More details are provided on the webpage of the HPC available at our host institution\footnote{The link will be revealed after the review process, to preserve blindness}.

Finally, as stated in the main paper, each of the experiments used $30$ CPUs and $4$GB of RAM.

\section{Description of the environments}
This section contains a description of the environments used for the experiments.
In the experimental setup, we used the following widely-known environments to test our method in a variety of conditions: small/large input spaces, small/(relatively) large action spaces, discrete/continuous action spaces, and types of tasks.

\subsection{InvertedPendulum-v2}
The InvertedPendulum-v2 task contains two main entities: a cart, which can freely move over the x-axis, and a pole, attached to the cart.
The goal of the task is to balance the pole for as long as possible by moving the cart.

The agent receives the following observations:
\begin{enumerate}
    \item $p_x$: position of the cart on the vertical axis
    \item $\theta$: angle between the vertical axis and the pole
    \item $v_x$: horizontal velocity of the cart
    \item $\omega$: angular velocity of the pole
\end{enumerate}

The agent has to control a single output variable: $f$, which is the force applied to the cart, in order to balance the pole.

The agent receives a reward of $1$ for each timestep in which the pole is balanced.

The simulation terminates whenever $\mid\theta\mid > 0.2$ or the simulation length exceeds $1000$ steps.

\subsection{LunarLander-v2}
The LunarLander-v2 environment simulates a lunar lander that must be controlled to land safely on the moon.

The agent receives an 8-dimensional observation containing:
\begin{enumerate}
\item $p_x, p_y$: coordinates of the lander
\item $v_x, v_y$: velocities of the lander
\item $\theta$: angle of the lander
\item $\omega$: angular velocity of the lander
\item $c_1, c_2$: Boolean values indicating whether each leg is in contact with the ground
\end{enumerate}
The agent can take 4 discrete actions:
\begin{enumerate}\addtocounter{enumi}{-1}
\item do nothing
\item fire left orientation engine
\item fire main engine
\item fire right orientation engine
\end{enumerate}
The agent receives a reward $r_t$ at each timestep based on:
\begin{itemize}
\item distance to the landing pad
\item lander speed
\item tilt angle
\item leg ground contact
\item engine firing
\end{itemize}
Additional rewards $r_{crash}$ and $r_{land}$ are given for crashing or landing safely.
The episode terminates if:
\begin{itemize}
\item the lander crashes
\item the lander moves outside the bounded area
\item the lander is inactive for too long
\end{itemize}

The goal is to land the lunar lander safely in the designated landing area. This environment is considered solved if the average score across 100 simulations is greater or equal to $200$ points.

\subsection{Swimmer-v2}
The Swimmer-v2 environment consists of a system with $3$ links connected by $2$ joints. The goal is to apply torques on the joints to move the swimmer to the right as quickly as possible.

The agent receives an observation containing:
\begin{enumerate}
\item $\theta^t$: angle of the tip of the body
\item $\theta^j_0$: angle of the first joint
\item $\theta^j_1$: angle of the second joint
\item $v^t_x, v^t_y$: linear velocities of the tip
\item $\omega^t$: angular velocity of the tip
\item $\omega^j_0, \omega^j_1$: angular velocities of the two joints
\end{enumerate}
The action space is defined in $[-1, 1]^2$, i.e., the torque for each of the joints.
The reward function has two components:
\begin{enumerate}
\item Forward reward: $0.04 \cdot (x^t_k+1 - x^t_k)$
\item Control cost: quadratic penalty on action magnitude, scaled by $10^{-4}$
\end{enumerate}

The episode terminates after 1000 timesteps.

\subsection{Reacher-v4}
In this task, the agent has to control a robotic manipulator with two joints. The goal is to move the end-effector as closely as possible to the target position. 

The observation space consists of 11 continuous variables:
\begin{enumerate}
    \item $cos(\theta_0)$: Cosine of the angle of the first arm.
    \item $cos(\theta_1)$: Cosine of the angle of the second arm.
    \item $sin(\theta_0)$: Sine of the angle of the first arm.
    \item $sin(\theta_1)$: Sine of the angle of the second arm.
    \item $p^t_x$: x-coordinate of the target position.
    \item $p^t_y$: y-coordinate of the target position.
    \item $\omega_0$: Angular velocity of the first arm.
    \item $\omega_1$: Angular velocity of the second arm.
    \item $d_x$: Distance between the end-effector and the target position on the x-axis
    \item $d_y$: Distance between the end-effector and the target position on the y-axis
    \item $d_z=0$: Distance between the end-effector and the target position on the z-axis, which is always zero
\end{enumerate}

All variables are continuous and unbounded.

The action space is defined in $[-1, 1]^2$. Each element of the vector represents the torques applied at the corresponding joint.
The reward function in the Reacher environment is composed of two parts:
\begin{itemize}
    \item $r_d$: Reward based on the distance, i.e. $-\lVert p^t - p^{ee}\rVert$, where $p^{ee}$ refers to the position of the end effector
    \item $r_c$: Penalizes large control values, and it is defined as the opposite of the sum of the action vector
\end{itemize}
An episode in the Reacher environment ends under the following conditions:
\begin{itemize}
    \item Truncation: The episode duration reaches 50 timesteps, with a new random target appearing if the Reacher's fingertip reaches it before 50 timesteps.
    \item Termination: Any of the state space values becomes non-finite.
\end{itemize}

\subsection{Hopper-v3}
The Hopper-v2 task simulates a body with $4$ main parts: a torso, thigh, leg, and foot connected by $3$ joints. The goal is to apply torques on the joints to make the hopper hop forward as far as possible without falling over.

The agent receives the following observations:
\begin{enumerate}
\item $p^t_z$: position of the torso w.r.t. the z-axis
\item $\theta^t$: angle of the torso
\item $\theta^{th}$: angle of the thigh
\item $\theta^{l}$: angle of the leg
\item $\theta^{f}$: angle of the foot
\item $v^{t}_x$: velocity of the torso w.r.t. the x-axis
\item $v^{t}_z$: velocity of the torso w.r.t. the z-axis
\item $\omega^{t}$: angular velocity of the torso
\item $\omega^{th}$: angular velocity of the thigh
\item $\omega^{l}$: angular velocity of the leg
\item $\omega^{f}$: angular velocity of the foot
\end{enumerate}

The agent has to control $3$ output variables $u_i \in [-1, 1]$, which are the torques applied to each joint $i$, in order to make the hopper hop forward.

The agent receives a reward $r_t$ at each timestep based on:
\begin{enumerate}
\item $r_h$: Healthy reward, equal to $1$ if hopper angles and height are within limits and $0$ otherwise
\item $r_f$: Forward reward for horizontal displacement equal to $0.008 \cdot (p^t_x(k+1) - p^t_x(k))$
\item $r_c$: Control cost penalty on action magnitude, equal to $0.001$ times the sum of the squares of the output variables.
\end{enumerate}

The simulation terminates whenever the hopper falls outside its limits or after 1000 timesteps.

\subsection{Walker2d}
The Walker2d-v2 task consists of a 2D bipedal walker with $7$ main parts: a torso, 2 thighs, 2 legs, and 2 feet, connected together by $6$ hinge joints. The goal is to apply torques to the joints to make the walker move forward as quickly as possible without falling over.

The agent receives observations containing:
\begin{enumerate}
\item $p_z$: z-position (height) of the torso
\item $\theta^t$: Angle of the torso
\item $\theta^{th_r}$: Angle of the right thigh
\item $\theta^{l_r}$: Angle of the right leg
\item $\theta^{f_r}$: Angle of the right foot
\item $\theta^{th_l}$: Angle of the left thigh
\item $\theta^{l_l}$: Angle of the left leg
\item $\theta^{f_l}$: Angle of the left foot
\item ${v}_x$: Velocity of the torso on the x-axis
\item ${v}_z$: Velocity of the torso on the z-axis
\item $\omega^{th_r}$: Angular velocity of the right thigh
\item $\omega^{l_r}$: Angular velocity of the right leg
\item $\omega^{f_r}$: Angular velocity of the right foot
\item $\omega^{th_l}$: Angular velocity of the left thigh
\item $\omega^{l_l}$: Angular velocity of the left leg
\item $\omega^{f_l}$: Angular velocity of the left foot
\end{enumerate}

The agent controls $6$ torque outputs $u_i \in [-1, 1]$, one for each joint $i$, to make the walker walk forward.

The reward $r_t$ consists of:
\begin{enumerate}
\item $r_h$: Healthy reward, equal to $1$ if hopper angles and height are within limits and $0$ otherwise
\item $r_f$: Forward reward for horizontal displacement equal to $0.008 \cdot (p^t_x(k+1) - p^t_x(k))$
\item $r_c$: Control cost penalty on action magnitude, equal to $0.001$ times the sum of the squares of the output variables.
\end{enumerate}

The episode terminates if the walker falls outside its healthy height and angle limits, or after 1000 timesteps.

\section{Social learning with neural networks}
The core concept underlying SIRL can be ported to other Reinforcement Learning techniques.
In fact, it is important to note that our approach is very convenient when there is a large number of agents to train, e.g., for hyperparameter optimization.

To showcase this possibility, here we present preliminary results for Social Learning with neural networks.
Here, as a proof of concept, we use Social Learning to reduce the computational cost of a grid search over the following hyperparameters:
\begin{itemize}[leftmargin=*]
 \item $\alpha \in \{10^{-3}, 10^{-4}\}$
 \item $n_l \in \{1, 3\}$
 \item $n_u \in \{8, 16\}$
 \item $b_s \in \{64, 128\}$
 \item $p \in \{100, 1000\}$
\end{itemize}
where $\alpha$ is the learning rate, $n_l$ is the number of layers for the DQN, $n_u$ is the number of units in each hidden layer, $b_s$ is the batch size, and $p$ is the target update period.

In this scenario, we initialize $32$ neural networks, one for each configuration of the grid search.
Then, we start the collaborative phase, training all of them simultaneously on $e_c = 500$ collaborative episodes.
Finally, each network is tested separately on $10$ additional episodes.

We test our Social Learning approach using Deep Q Learning (using Deep Q Networks, DQN) on the CartPole-v1 task.
\Cref{fig:deep_social} shows the cumulative rewards obtained by the agent during the training, compared to that of the best configuration from the grid search, trained from scratch in a non-social fashion (i.e., traditional Deep Q Learning).
Interestingly, we can observe that up to the $500$-th episode, the scores obtained by all the agents trained in a social manner are comparable to those of the DQN trained alone.
However, when switching to the individual phase, the scores obtained by the same network trained with social learning are significantly higher than those of the non-social DQN.

Even though these results represent just a preliminary experimentation with neural networks and Social Learning, it seems that Social Learning gives a boost over traditional Deep Q Learning.
In fact, the agents achieve significantly better performance, while reducing the computational cost of the grid search by $95\%$, as shown graphically in \Cref{fig:deep_barplot}.

\begin{figure}[!ht]
 \begin{subfigure}[b]{0.48\textwidth}
 \centering
 \begin{tikzpicture}[/pgfplots/set layers, spy using outlines={circle, magnification=3, size=2cm, connect spies}]
\begin{axis}[
    xlabel={Episode},
    axis on top,
    ylabel={Cumulative Reward},
    legend style={at={(0.5,-0.25)},anchor=north},
    ymin=-0, ymax=500, 
    xmin=-0, xmax=510, 
    ytick={0, 100, 200, 300, 400, 500},
    cycle list={blue,red}, 
    title={CartPole-v1},
    legend cell align={left},
    axis background/.style={fill=gray!20},
    grid=major,
    grid style={white},
    major grid style={line width=1pt},
    width=\textwidth,
    ]
    \spy[black,size=1cm,fill=none,rectangle] on (0.8\textwidth,0.18\textwidth) in node at (0.8,2.5);
    \spy[black,size=1cm,fill=none,rectangle] on (0.8\textwidth,0.65\textwidth) in node at (2,2.5);
    \linewitherrordifferent{imgs/NNs/summ_CartPole-v1_500_0.csv}{Episode}{Mean}{LowCI}{blue}
    \linewitherrordifferent{imgs/NNs/summ_CartPole-v1_500_1.csv}{Episode}{Mean}{LowCI}{red}
    \legend{DQN, Social DQN}
\end{axis}
\end{tikzpicture}
 \caption{}
 \label{fig:deep_social}
 \end{subfigure}
 \hfill
 \begin{subfigure}[b]{0.48\textwidth}
 \centering
 \begin{tikzpicture}

\begin{axis} [
    xbar = .05cm,
    bar width = 10pt,
    xmin = 0, 
    xmax = 20000, 
    enlarge y limits = {abs = .5},
    enlarge x limits = {value = 0, upper},
    width = \textwidth,
    height = 5.8cm,
    cycle list={green,red,blue}, 
    ytick={0, 1, 2},
    yticklabels={CartPole-v1},
    legend image code/.code={
        \draw [#1] (0cm,-0.1cm) rectangle (0.2cm,0.25cm);},
    legend cell align={left},
    xlabel = {Episodes ($\times 10^4$)},
    xtick pos=left,
    ytick pos=left,
    xtick scale label code/.code={},
    legend style={at={(0.5,-0.25)},anchor=north},
    reverse legend,
    yticklabel style={rotate=90}
]
\legend{Social DQN, DQN}
\addplot[fill=red!50!white] coordinates {(820,0)};
\addplot[fill=blue!50!white] coordinates {(16320,0)};
\end{axis}

\end{tikzpicture}
 \vspace{8pt}
 \caption{}
 \label{fig:deep_barplot}
 \end{subfigure}
 \caption{Results obtained by applying our social learning approach to hyperparameter optimization for Deep RL. (a) Scores obtained by the best neural network found from the grid search when trained with Deep Q Learning and Social Deep Q Learning. The solid lines represent the mean over $10$ independent runs, while the shaded area represents the $95\%$ CI. (b) Number of episodes simulated with the environment by two different versions of the grid search. ``DQN'' refers to a grid search using traditional Deep Q learning, while ``Social DQN'' refers to a grid search using our social learning approach.}
 \label{fig:deep_sirl}
\end{figure}

\section{Parallelization}
\label{sec:parallelization}
The parallelization scheme is shown in \Cref{fig:parallel}.
To parallelize the collaborative phase, we can divide the number of collaborative episodes in $e_c = e_p \cdot i$, where $e_p$ is the number of parallel episodes, and $i$ is the number of episodes simulated by each parallel process.

Thus, at the beginning of the collaborative phase, a copy of each DT is sent to each parallel job, where it learns using Q-learning.
After all the parallel jobs are terminated, each tree $T_i$ is updated as the \emph{average of all the trees} trained in the parallel processes $T_{i,1}, \dots, T_{i, e_p}$.
We define an average tree as a tree that has, for all the Q-values of each leaf, the average across all the $e_p$ trees trained in parallel.

It is worth noting that we can use this parallelization scheme since:
\begin{enumerate}[leftmargin=*]
 \item The Q-values contained on the leaves are constants, so there are no non-linear operations involved in their computation;
 \item The learning rate $\alpha$ is fixed;
 \item The splits of each tree are kept constant throughout the evaluation phase.
\end{enumerate}

\tikzset{
dot/.style = {circle, fill, minimum size=#1,
 inner sep=0pt, outer sep=0pt},
dot/.default = 6pt 
}

\begin{figure*}[!ht]
 \centering
 \begin{tikzpicture}
 \node[rectangle, draw=cbblue, fill=cbblue!10!white, rounded corners, thick, minimum width=\textwidth, minimum height=5.5cm, text depth=5.5cm] (coll) at (1.85, 0) {Collaborative phase};
 \node[rectangle, draw, rounded corners, minimum height=4cm, minimum width=1cm] at (-4.5,0) {};
 \node[] (t1) at (-4.5,1) {\large$T_1$};
 \node[] (t2) at (-4.5,0) {\large$T_2$};
 \node[] (t3) at (-4.5,-1) {\large$T_3$};
 \node[rectangle, draw=cbblue, fill=cbblue!10!white, rounded corners, thick, minimum width=0.5\textwidth, minimum height=1cm] (coll1) at (1, 2) {$i$ episodes with $T_{1,1}, T_{2,1}, T_{3,1}$};
 \node[rectangle, draw=cbblue, fill=cbblue!10!white, rounded corners, thick, minimum width=0.5\textwidth, minimum height=1cm] (coll2) at (1, 0.5) {$i$ episodes with $T_{1,2}, T_{2,2}, T_{3,2}$};
 \node[rectangle, draw=cbblue, fill=cbblue!10!white, rounded corners, thick, minimum width=0.5\textwidth, minimum height=1cm] (coll3) at (1, -2) {$i$ episodes with $T_{1,e_p}, T_{2,e_p}, T_{3,e_p}$};
 \draw [decorate, decoration = {brace}] (0.8,-1.25) -- (0.8,-0.25);
 \node at (0., -0.75) {\shortstack{$e_p$ \\ parallel \\ processes}};
 \node[dot] at (1,-0.25) {};
 \node[dot] at (1,-0.75) {};
 \node[dot] at (1,-1.25) {};
 \node[rectangle, draw, rounded corners, minimum height=4cm, minimum width=2.5cm] at (7.4,0) {};
 \node[] (t1n) at (7.4,1) {\large$T_1 = \underset{i}{\mathbb{E}}[T_{1,i}]$};
 \node[] (t2n) at (7.4,0) {\large$T_2 = \underset{i}{\mathbb{E}}[T_{2,i}]$};
 \node[] (t3n) at (7.4,-1) {\large$T_3 = \underset{i}{\mathbb{E}}[T_{3,i}]$};
 \draw[->] (t1) -- (coll1.west);
 \draw[->] (t1) -- (coll2.west);
 \draw[->] (t1) -- (coll3.west);
 \draw[->] (t2) -- (coll1.west);
 \draw[->] (t2) -- (coll2.west);
 \draw[->] (t2) -- (coll3.west);
 \draw[->] (t3) -- (coll1.west);
 \draw[->] (t3) -- (coll2.west);
 \draw[->] (t3) -- (coll3.west);
 \draw[->] (coll1.east) -- (t1n.west);
 \draw[->] (coll1.east) -- (t2n.west);
 \draw[->] (coll1.east) -- (t3n.west);
 \draw[->] (coll2.east) -- (t1n.west);
 \draw[->] (coll2.east) -- (t2n.west);
 \draw[->] (coll2.east) -- (t3n.west);
 \draw[->] (coll3.east) -- (t1n.west);
 \draw[->] (coll3.east) -- (t2n.west);
 \draw[->] (coll3.east) -- (t3n.west);
 \end{tikzpicture}
 \caption{Parallelization scheme for the collaborative phase.}
 \label{fig:parallel}
\end{figure*}

\section{Best decision trees}
In this section, we show the best (pruned) DTs that we obtained for each of the tasks.
To prune the DTs, we remove all the nodes that, in $100$ episodes, have been visited less than the $0.5\%$ of the visits received by their parent.
More specifically, for each node (except the root), we compute the ratio of visits that the node $n$ received: $r_n = \frac{v_n}{v_p}$, where $p$ is the parent of $n$.
If $r_n < 0.005$, we conclude that the node is of marginal importance and thus it is not needed.
Therefore, we remove the node $n$ and connect the parent of $p$ with the sibling of $n$, so that the parent node is bypassed.

\subsection{InvertedPendulum-v2}
The best DT for InvertedPendulum-v2 is extremely simple, and it is shown in \Cref{fig:best_ip}.
The weight for the linear combination is $\textit{w}_1 = [-0.224, -5.024, -0.684, -1.678]$ and the constant used for the comparison (i.e., the bias) is $b_1 = 0.285$.

\begin{figure}[!ht]
 \centering
 \begin{tikzpicture}
 \node[condition, ] (root) {$\mathbf{w}_1^T\mathbf{x} < b_1$};
 \node[leaf, below left = of root] (rootl) {0.67};
 \node[leaf, below right = of root] (rootr) {-1};
 \draw (root) -| (rootl);
 \draw (root) -| (rootr);
 \end{tikzpicture}
 \caption{Best DT found for the InvertedPendulum-v2 environment, after pruning.}
 \label{fig:best_ip}
\end{figure}

\subsubsection{Interpretation}
The interpretation of this DT is quite straightforward.
The InvertedPendulum-v2 environment provides the following variables as input:
\begin{itemize}[leftmargin=*]
 \item $x$: the linear position of the cart
 \item $\theta$: the angular position of the cart
 \item $v$: the linear velocity of the cart
 \item $\omega$: the angular velocity of the cart
\end{itemize}

The action of the agent is a continuous value in $[-1, 1]$, that represents the power used by the agent to accelerate towards the left (negative values) or the right (positive values).

The agent receives a reward of $1$ for each step that the pole is upright.

Thus, we can write the condition of the DT as follows:
\begin{equation}
 -0.224x -5.024\theta -0.684v -1.678\omega < 0.285
\end{equation}

Grouping the terms related to the linear part and to the angular part, we get:
\begin{equation}
 \alpha (x + k_v v) + \beta (\theta + k_\omega \omega) < t 
\end{equation}

Noting that $x_k = x_{k-1} + T v_{k-1}$ and $\theta_k = \theta_{k-1} + T \omega_{k-1}$, where $k$ refers to the current step, we hypothesize that the system works as follow:
it tries to compute a weighted sum of the estimated future position and estimated future angular velocity.
Then, if that sum is unbalanced to the right, it accelerates to the left and vice versa.

However, it is worth noting that the coefficients are not used in such a way that $k_v = k_\omega = T = 0.02$, as shown in the documentation of the MuJoCo library\footnote{\url{https://github.com/openai/gym/blob/dcd185843a62953e27c2d54dc8c2d647d604b635/gym/envs/mujoco/assets/inverted\_pendulum.xml}}.
In fact, we hypothesize that $k_v$ and $k_\omega$ are much greater than $T$ in order to lead to a \emph{worst-case} estimation of the future positions and velocities, to have a stable policy.

To test this hypothesis, we measure the cumulative reward of our approach when using different values for $k_v$ and $k_\omega$. The results are shown in \Cref{fig:rob}.
We can observe, in fact, that the values used (i.e., the ones represented by dashed lines) reflect indeed parameters that have the highest score with the smallest standard deviation.
Moreover, it is interesting to note that, for the $k_v$ parameter, all the values above a certain value (i.e., about $2.5$) lead to solutions with a very good score.
Instead, for the $k_\omega$ parameter, there is a relatively narrow window that allows for an overestimation of the future angle that is good enough to solve the task.

\begin{figure}[!ht]
 \centering
 \begin{subfigure}{0.45\textwidth}
 \begin{tikzpicture}[/pgfplots/set layers]
\begin{axis}[
    xlabel={$k_v$},
    axis on top,
    ylabel={Cumulative Reward},
    legend style={at={(0.5,-0.2)},anchor=north},
    ymin=-0, ymax=1100, 
    xmin=-0, xmax=4.7, 
    cycle list={blue,red,green}, 
    title={InvertedPendulum-v2},
    width=\textwidth,
    xtick pos=left,
    ytick pos=left,
    ]
    
    \linewitherrordifferent{./imgs/DTs/test_robustness.csv}{vM}{return}{std}{blue}
    \draw [dashed, color=black] (3,0) -- (3,1000);
\end{axis}
\end{tikzpicture}
 \end{subfigure}
 \hfill
 \begin{subfigure}{0.45\textwidth}
 \begin{tikzpicture}[/pgfplots/set layers]
\begin{axis}[
    xlabel={$k_\omega$},
    axis on top,
    ylabel={Cumulative Reward},
    legend style={at={(0.5,-0.2)},anchor=north},
    ymin=-0, ymax=1100, 
    xmin=-0, xmax=1, 
    cycle list={blue,red,green}, 
    title={InvertedPendulum-v2},
    width=\textwidth,
    xtick pos=left,
    ytick pos=left,
    ]
    
    \linewitherrordifferent{./imgs/DTs/test_wm.csv}{wM}{return}{std}{red}
    \draw [dashed, color=black] (0.33,0) -- (0.33,1000);
\end{axis}
\end{tikzpicture}
 \end{subfigure}
 \caption{Cumulative reward obtained by varying the $k_v$ and $k_\omega$ coefficients. The solid blue line represents the mean over $100$ episodes, while the shaded area represents the standard deviation. The dashed line represents the values from the original solution.}
 \label{fig:rob}
\end{figure}

\subsection{LunarLander-v2}
The best DT for this environment is shown in \Cref{fig:best_ll}.
The weights and the biases for each of the oblique splits are shown below.

The weights for the linear combinations of the inputs are:
\begin{itemize}[leftmargin=*]
\item $ \mathbf{w}_1 = [-7.2, -4.4, -7.9, -4.7, 6.7, -9.5, 8.7, -3.7]$
\item $ \mathbf{w}_2 = [-6.3, 0.8, -5.4, 4.5, 5.7, 5.7, 1.4, -1.2]$
\item $ \mathbf{w}_3 = [4.4, 0.9, 6.8, 4.5, -9.1, -9.4, 7.3, 8.2]$
\item $ \mathbf{w}_4 = [5.6, 1.5, 4.7, -1.2, -7.7, -9.1, 7.3, 7.9]$
\item $ \mathbf{w}_5 = [3.9, -9.6, -5.8, -2.6, -3.1, 4.8, 2.7, -7.8]$
\end{itemize}

The biases (i.e., constants) for the splits are:
\begin{itemize}[leftmargin=*]
\item $ b_1 = 7.072$
\item $ b_2 = -0.955$
\item $ b_3 = -0.606$
\item $ b_4 = 9.66$
\item $ b_5 = 7.074$
\end{itemize}

\begin{figure}[!ht]
 \centering
 \resizebox{0.9\columnwidth}{!}{
 \begin{tikzpicture}
 \node[condition, ] (root) {$\mathbf{w}_1^T\mathbf{x} < b_1$};
 \node[oblique=root/rootl/2/left] {};
 \node[oblique=rootl/rootll/3/left, xshift=-1cm] {};
 \node[leaf, below left = of rootll] (rootlll) {2};
 \node[leaf, below right = of rootll] (rootllr) {1};
 \draw (rootll) -| node[midway, above] {T} (rootlll);
 \draw (rootll) -| node[midway, above] {F} (rootllr);
 \node[oblique=rootl/rootlrr/4/right, xshift=1cm] {};
 \node[oblique=rootlrr/rootlrrl/5/left, yshift=-0.5cm] {};
 \node[leaf, below left = of rootlrrl] (rootlrrll) {3};
 \node[leaf, below right = of rootlrrl] (rootlrrlr) {2};
 \draw (rootlrrl) -| node[midway, above] {T} (rootlrrll);
 \draw (rootlrrl) -| node[midway, above] {F} (rootlrrlr);
 \node[leaf, below right = of rootlrr] (rootlrrr) {0};
 \draw (rootlrr) -| node[midway, above] {T} (rootlrrl);
 \draw (rootlrr) -| node[midway, above] {F} (rootlrrr);
 \draw (rootl) -| node[midway, above] {T} (rootll);
 \draw (rootl) -| node[midway, above] {F} (rootlrr);
 \node[leaf, below right = of root] (rootr) {2};
 \draw (root) -| node[midway, above] {T} (rootl);
 \draw (root) -| node[midway, above] {F} (rootr);
 \end{tikzpicture}
 }
 \caption{Best DT found for the LunarLander-v2 environment, after pruning.}
 \label{fig:best_ll}
\end{figure}

\subsubsection{Interpretation}
Since this tree is deeper, we will perform a higher-level interpretation of the solution w.r.t. the one provided for InvertedPendulum-v2.

The observation from the environment consists of the following variables: 
\begin{itemize}[leftmargin=*]
 \item $p_x$: Horizontal position
 \item $p_y$: Vertical position
 \item $v_x$: Horizontal velocity
 \item $v_y$: Vertical velocity
 \item $\theta$: Angle w.r.t. vertical axis
 \item $\omega$: Angular velocity
 \item $c_l$: Boolean indicating contact between the left leg and the ground
 \item $c_r$: Boolean indicating contact between the right leg and the ground
\end{itemize}

The actions that the agents can perform are:
\begin{itemize}[leftmargin=*]
    \item $0$: NOP, all the engines are shut down
    \item $1$: Right engine, pushes the lander to the left
    \item $2$: Main engine, accelerates the lander upward
    \item $3$: Left engine, pushes the lander to the right
\end{itemize}

The reward obtained by the agent depends on the positions and velocities (both linear and angular) of the lander.
Moreover, there is a bonus for touching the ground with the legs; and a malus for firing the engines.

We can easily deduce that the policy is biased towards action $2$ (i.e., main engine).
In fact, it makes use of condition $w_3$ to detect, using the right branch, if the lander is falling to the right. If so, it executes action $1$, i.e. fire the right engine.
On the other hand, it uses condition $w_5$ (left branch) to check whether the lander is falling to the left. If so, it executes the left engine, to balance the lander.
Moreover, by observing the policy while simulating its behavior in the environment, we deduce that action $0$ (i.e., NOP) is performed when the lander has finished the landing process.

\subsection{Swimmer-v2}
The best DT found for the Swimmer-v2 environment is shown in \Cref{fig:best_sw}.
This DT is significantly bigger than those seen previously.
This is likely due to the fact that, by using the discretization of the action space shown in the ``Hyperparameters'' section, the algorithm has to find a DT that, at each step, determines what is the most important actuator to use for the current action.
We hypothesize that using different methods for discretizing the action space could lead to smaller DTs.
In fact, if the discretization is made in such a way that the actions are not decoupled, it may be possible that the optimization process will discover simpler trees.
However, such an approach would increase the number of episodes needed to thoroughly explore the action space.

The weights for the linear combinations of the inputs are:
\begin{itemize}[leftmargin=*]
 \item $ \mathbf{w}_{1} = [-6.8, -0.2, 7.6, 4.3, -7.7, 3.4, 9.8, 1.1]$
 \item $ \mathbf{w}_{2} = [-8.4, 7.0, 3.0, 1.0, 1.5, 0.1, 7.5, -0.2]$
 \item $ \mathbf{w}_{3} = [-9.0, 5.7, -7.1, -9.9, 4.9, 2.9, -1.8, -6.6]$
 \item $ \mathbf{w}_{4} = [3.7, -4.1, -3.0, 4.3, -1.0, 3.7, -0.3, 2.5]$
 \item $ \mathbf{w}_{5} = [-0.4, 8.2, -9.3, -7.0, -8.2, -6.6, -8.1, -9.5]$
 \item $ \mathbf{w}_{6} = [6.3, 2.1, 1.0, 6.9, -6.0, -6.6, -7.1, -1.7]$
 \item $ \mathbf{w}_{7} = [6.2, 3.3, -3.6, -9.0, -2.1, -6.5, -8.0, 1.0]$
 \item $ \mathbf{w}_{8} = [-4.4, 5.8, -2.9, -0.1, -8.9, 4.5, -9.0, -5.4]$
 \item $ \mathbf{w}_{9} = [1.4, -3.5, -8.3, 6.3, 9.0, 7.8, -0.4, -5.1]$
 \item $ \mathbf{w}_{10} = [-1.2, 2.3, 3.2, -7.3, 1.8, -7.5, 4.4, 6.1]$
 \item $ \mathbf{w}_{11} = [-7.8, 8.0, -0.5, -5.4, 7.3, -5.2, 7.2, -1.5]$
 \item $ \mathbf{w}_{12} = [4.2, -6.7, -7.4, -5.5, 1.6, -1.1, -8.9, 5.7]$
 \item $ \mathbf{w}_{13} = [3.6, -2.4, -0.7, 4.4, -2.4, 0.0, 0.0, 0.0]$
\end{itemize}

The biases (i.e., constants) for the splits are:
\begin{itemize}[leftmargin=*]
 \item $ b_{1} = -3.519$
 \item $ b_{2} = 4.132$
 \item $ b_{3} = 8.252$
 \item $ b_{4} = 3.955$
 \item $ b_{5} = -6.651$
 \item $ b_{6} = 0.115$
 \item $ b_{7} = 7.284$
 \item $ b_{8} = 2.605$
 \item $ b_{9} = 1.749$
 \item $ b_{10} = -1.853$
 \item $ b_{11} = -7.113$
 \item $ b_{12} = -1.611$
 \item $ b_{13} = 0.0$
\end{itemize}

\begin{figure}[!ht]
 \centering
 \begin{tikzpicture}[node distance=0.5cm]
 \node[condition, ] (root) {$\mathbf{w}_1^T\mathbf{x} < b_1$};
 \node[leaf, below left = of root] (rootl) {$[0.67, 0]$};
 \node[oblique=root/rootr/2/right, yshift=0cm] {};
 \node[oblique=rootr/rootrl/3/left, yshift=0cm] {};
 \node[oblique=rootrl/rootrll/4/left, yshift=0cm] {};
 \node[oblique=rootrll/rootrlll/5/left, yshift=0cm] {};
 \node[leaf, below left = of rootrlll] (rootrllll) {$[-0.67, 0]$};
 \node[oblique=rootrlll/rootrlllr/6/right, yshift=0cm] {};
 \node[leaf, below left = of rootrlllr] (rootrlllrl) {$[-0.33, 0]$};
 \node[oblique=rootrlllr/rootrlllrr/7/right, yshift=0cm] {};
 \node[oblique=rootrlllrr/rootrlllrrl/8/left, yshift=0cm] {};
 \node[leaf, below left = of rootrlllrrl] (rootrlllrrll) {$[-0.33, 0]$};
 \node[oblique=rootrlllrrl/rootrlllrrlr/9/right, yshift=0cm] {};
 \node[oblique=rootrlllrrlr/rootrlllrrlrl/10/left, yshift=0cm] {};
 \node[oblique=rootrlllrrlrl/rootrlllrrlrll/11/left, yshift=0cm] {};
 \node[leaf, below left = of rootrlllrrlrll] (rootrlllrrlrlll) {$[-1, 0]$};
 \node[oblique=rootrlllrrlrll/rootrlllrrlrllr/12/right, yshift=0cm] {};
 \node[oblique=rootrlllrrlrllr/rootrlllrrlrllrl/13/left, yshift=0cm] {};
 \node[leaf, below left = of rootrlllrrlrllrl] (rootrlllrrlrllrll) {$[0, 1]$};
 \node[leaf, below right = of rootrlllrrlrllrl] (rootrlllrrlrllrlr) {$[1, 0]$};
 \draw (rootrlllrrlrllrl) -| (rootrlllrrlrllrll);
 \draw (rootrlllrrlrllrl) -| (rootrlllrrlrllrlr);
 \node[leaf, below right = of rootrlllrrlrllr] (rootrlllrrlrllrr) {$[1, 0]$};
 \draw (rootrlllrrlrllr) -| (rootrlllrrlrllrl);
 \draw (rootrlllrrlrllr) -| (rootrlllrrlrllrr);
 \draw (rootrlllrrlrll) -| (rootrlllrrlrlll);
 \draw (rootrlllrrlrll) -| (rootrlllrrlrllr);
 \node[leaf, below right = of rootrlllrrlrl] (rootrlllrrlrlr) {$[-0.33, 0]$};
 \draw (rootrlllrrlrl) -| (rootrlllrrlrll);
 \draw (rootrlllrrlrl) -| (rootrlllrrlrlr);
 \node[leaf, below right = of rootrlllrrlr] (rootrlllrrlrr) {$[-0.67, 0]$};
 \draw (rootrlllrrlr) -| (rootrlllrrlrl);
 \draw (rootrlllrrlr) -| (rootrlllrrlrr);
 \draw (rootrlllrrl) -| (rootrlllrrll);
 \draw (rootrlllrrl) -| (rootrlllrrlr);
 \node[leaf, below right = of rootrlllrr] (rootrlllrrr) {$[0, -0.67]$};
 \draw (rootrlllrr) -| (rootrlllrrl);
 \draw (rootrlllrr) -| (rootrlllrrr);
 \draw (rootrlllr) -| (rootrlllrl);
 \draw (rootrlllr) -| (rootrlllrr);
 \draw (rootrlll) -| (rootrllll);
 \draw (rootrlll) -| (rootrlllr);
 \node[leaf, below right = of rootrll] (rootrllr) {$[0, -1]$};
 \draw (rootrll) -| (rootrlll);
 \draw (rootrll) -| (rootrllr);
 \node[leaf, below right = of rootrl] (rootrlr) {$[-0.67, 0]$};
 \draw (rootrl) -| (rootrll);
 \draw (rootrl) -| (rootrlr);
 \node[leaf, below right = of rootr] (rootrr) {$[0, 1]$};
 \draw (rootr) -| (rootrl);
 \draw (rootr) -| (rootrr);
 \draw (root) -| (rootl);
 \draw (root) -| (rootr);
 \end{tikzpicture}
 \caption{Best DT found for the Swimmer-v2 environment, after pruning.}
 \label{fig:best_sw}
\end{figure}

\subsubsection{Interpretation}
The inputs provided by this environment are the following:
\begin{itemize}[leftmargin=*]
    \item $\theta_0$: angle of the front tip
    \item $\theta_1$: angle of the first rotor
    \item $\theta_2$: angle of the second rotor
    \item $v_x$: velocity of the tip along the x-axis
    \item $v_y$: velocity of the tip along the y-axis
    \item $\omega_0$: angular velocity of front tip
    \item $\omega_1$: angular velocity of first rotor
    \item $\omega_2$: angular velocity of second rotor
\end{itemize}

The agent needs to control the torque applied to the two joints of the body.
This means that an action of the agent is $\mathbf{a}: [-1, 1]^2$, whose values represent the normalized torque applied to each of the joints.

Finally, the reward is computed by performing a weighted sum of the distance traveled by the agent and the cost of control, i.e., the $\mathcal{L}^2$ norm of the action.

The DT obtained in this environment is rather complex.
However, we can easily observe that this DT is collapsed to a decision list (i.e., it is extremely unbalanced).
This means that it can be seen as a set of consecutive checks to be performed.

While complicated, we can quickly deduce that this policy tries to minimize the control cost.
In fact, while most of the nodes on the top do not use the maximum values for the actions, the ones on the bottom (i.e., $w_{11}, w_{12}, w_{13}$) use the maximum power.
This means that the policy tries to check whether it can move the robot with a fraction of the maximum power. If all the attempts fail, then it uses the maximum power to move the robot.

\subsection{Reacher-v4}
The tree found in this environment (shown in \Cref{fig:best_re}) is quite trivial.
In fact, given the weights 
$w_1 = [0.3, 9.3, 0.0, 0.0, -9.8, 5.1, 0.2, 1.1, -5.1, 0.0, 0.0]$ and the bias $b_1 = 0$, we can conclude that the policy works as follows.
\[ 0.3 cos(\theta_0) + 9.3 cos(\theta_1) - 9.8 p^t_x + 5.1 p^t_y + 0.2 \omega_0 + 1.1 \omega_1 - 5.1 d_x\]

\subsubsection{Interpretation}
By analyzing the magnitudes of the variables, together with their multipliers shown above, we observe that, with a small loss in performance, we can simplify the main condition of the policy to:
\[ 9.3 cos(\theta_1) + 1.1 \omega_1 < 0\]
This is consistent with the fact that the policy only applies torques to the second joint, ignoring the first one.

Thus, when $\theta_1 \geq 0.12\omega_1$, the agent applies a torque of $-0.33 N\cdot m$ to the second joint, otherwise it does not apply any torque.

\begin{figure}[!ht]
 \centering
 \begin{tikzpicture}[node distance=0.5cm]
 \node[condition, ] (root) {$\mathbf{w}_1^T\mathbf{x} < b_1$};
 \node[leaf, below left = of root] (rootl) {$[0.00, 0.00]$};
 \node[leaf, below right = of root] (rootr) {$[0.00, -0.33]$};
 \draw (root) -| (rootl);
 \draw (root) -| (rootr);
 \end{tikzpicture}
 \caption{Best DT found for the Reacher-v4 environment, after pruning.}
 \label{fig:best_re}
\end{figure}

\subsection{Hopper-v3}
The best DT for this environment is shown in \Cref{fig:best_hop}.
The weights used by this DT are the following:
\begin{itemize}
    \item $w_1 = [0, 10, -7.4, 0, 6.9, 6.9, 0, 0, 0, 0, 0]$
    \item $w_2 = [0, 10, 0, 0, -6.8, 0, -6, 0, -4.2, 4.2, -7.4]$
    \item $w_3 = [0, 3.8, 0, 0, 5.2, 0, -8.2, 3.1, 0, -4.3, 0]$
\end{itemize}
while the biases are:
\begin{itemize}
    \item $b_1 = 3.1$
    \item $b_2 = 1.5$
    \item $b_3 = -0.5$
\end{itemize}

\subsubsection{Interpretation}
This DT appears to have found a rather simple policy for this environment, using only three nodes to decompose the states of the environment.

Moreover, it is important to note that this policy only balances the system, without making it actually ``hop''.
We hypothesize that this is caused by our discretization method, which does not allow to have complex control policies.
Thus, the agent only tries to exploit the ``healthy reward'' by not making the system fall.

The first condition can be rewritten as:
\[10\theta^t-7.4\theta^{th}+6.9\theta^f+6.9v^t_x < 3.1,\]
from numerical experiments, it emerged that the term $-7.4\theta^{th}$ has a negligible impact on the policy.
Thus, we can refactor the condition as:
\[10\theta^t+6.9\theta^f+6.9v^t_x < 3.1.\]
From \Cref{fig:best_hop}, we observe that, if the condition evaluates to false, i.e.:
\[6.9\theta^f \geq 3.1-10\theta^t-6.9v^t_x,\]
it is equivalent to say that
\[\theta^f \geq c_1-c_2\theta^t-v^t_x.\]
Thus, we can say that, when this condition evaluates to false, the agent rotates the foot towards the left to compensate the foot angle and the rest of the body by decreasing the foot angle.

The second condition can be rewritten as:
\[10\theta^t-6.8\theta^f-6v^t_z-4.2\omega^{th}+\omega^l-7.4\omega^f < 1.5\]

Also in this case, we conducted experiments to study the impact of each term on the policy. From these experiments, it emerged that we can safely reduce the policy to:
\[-6v^t_z-7.4\omega^f < 1.5\]
Thus, we hypothesize that when this condition evaluates to true, i.e.,
\[\omega^f > c_3 - c_4 v^t_z \], the agent tries to balance the foot position by tilting the leg.
Moreover, additional experiments showed that changing the magnitude of the force exerted on the leg joint does not significantly change the outcome, as long as the force is positive (for better precision, we found that, using a value of $0.01$ instead of $0.67$ reduced the average score by about $300$ points, without significantly affecting the behavior of the policy).

Finally, the last condition can be written as:
\[3.8\theta^t+5.2\theta^f-8.2v^t_z+3.1\omega^t-4.3\omega^l < -0.5\]

From numerical experiments, we observed that several terms can be removed, reducing the condition as:
\[3.1\omega^t - 4.3\omega^l < -0.5,\]
which can be rewritten as:
\[\omega^t < c_5\omega^l - c_6.\]
This can be interpreted as follows.
If the angle of the torso is (about) lower than the angle of the leg minus some constant, then try to compensate and apply a positive torque to the thigh.
Otherwise, apply a torque to the foot to try to keep the ``robot'' standing.
To better validate this hypothesis, we study the role of the constants in this condition.
The results of the sensitivity analysis are shown in \Cref{fig:sa_hop_c4} for $c_4$ and in \Cref{fig:sa_hop_c5} for $c_5$.
We observe that, while the $c_4$ parameter has a minor impact on the cumulative reward, the threshold $c_5$ can have a major effect on it.
In fact, \Cref{fig:sa_hop_c5} shows that all the negative values approximately lead to the same cumulative reward.
On the other hand, positive values completely disrupt the policy, making it significantly less performing.
This allows us to hypothesize that this is due to the fact that the condition evaluates to ``true'' rarely, but its role is crucial, as it allows balancing the system to prevent its falling.

\begin{figure}[!ht]
    \centering
    \resizebox{0.9\columnwidth}{!}{
    \begin{tikzpicture}
        \node[condition, ] (root) {$\mathbf{w}_1^T\mathbf{x} < b_1$};
        \node[oblique=root/rootl/2/left, xshift=0cm]{};
        \node[leaf, below left = of rootl] (rootll){[0, 0.67, 0]};
        \node[oblique=rootl/rootlr/3/right, xshift=0cm]{};
        \node[leaf, below left = of rootlr] (rootlrl){[1.0, 0, 0]};
        \node[leaf, below right = of rootlr] (rootlrr){[0, 0, 1.0]};
        \draw (rootlr) -| node[midway, above]{T}(rootlrl);
        \draw (rootlr) -| node[midway, above]{F}(rootlrr);
        \draw (rootl) -| node[midway, above]{T}(rootll);
        \draw (rootl) -| node[midway, above]{F}(rootlr);
        \node[leaf, below right = of root] (rootr){[0, 0, -0.67]};
        \draw (root) -| node[midway, above]{T}(rootl);
        \draw (root) -| node[midway, above]{F}(rootr);

    \end{tikzpicture}
    }
    \caption{Best DT found for the Hopper-v3 environment.}
    \label{fig:best_hop}
\end{figure}

\begin{figure}
    \centering
    \begin{tikzpicture}[/pgfplots/set layers]
\begin{axis}[
    xlabel={$c_4$},
    axis on top,
    ylabel={Cumulative Reward},
    legend style={at={(0.5,-0.2)},anchor=north},
    ymin=0, ymax=1100, 
    xmin=-4, xmax=4, 
    cycle list={blue,red,green}, 
    title={Hopper-v3},
    width=0.9\columnwidth,
    xtick pos=left,
    ytick pos=left,
    ]
    
    \linewitherrordifferent{./imgs/DTs/c4_params.csv}{c4}{mean}{std}{blue}
\end{axis}
\end{tikzpicture}
    \caption{Sensitivity analysis for the $c_4$ parameter of the best DT trained on Hopper-v3}
    \label{fig:sa_hop_c4}
\end{figure}

\begin{figure}
    \centering
    \begin{tikzpicture}[/pgfplots/set layers]
\begin{axis}[
    xlabel={$c_5$},
    axis on top,
    ylabel={Cumulative Reward},
    legend style={at={(0.5,-0.2)},anchor=north},
    ymin=-0, ymax=1100, 
    xmin=-4, xmax=4, 
    cycle list={blue,red,green}, 
    title={Hopper-v3},
    width=0.9\columnwidth,
    xtick pos=left,
    ytick pos=left,
    ]
    
    \linewitherrordifferent{./imgs/DTs/c5_params.csv}{c5}{mean}{std}{red}
\end{axis}
\end{tikzpicture}
    \caption{Sensitivity analysis for the $c_5$ parameter of the best DT trained on Hopper-v3}
    \label{fig:sa_hop_c5}
\end{figure}

\subsection{Walker2d-v3}
The best DT found for the Walker2d-v3 environment is shown in \Cref{fig:best_w2d}.
Its weights are:
\begin{itemize}
    \item $w_1 =[0.9, 8.4, 0, 5.9, 0, 0, 2.6, 5.7, -0.8, 5.7, 7.7, 0, -8, \\ -0.9, -6.8, 0, 3.3]$
    \item $w_2 = [0, 0, 0, -9.2, 0, 0, 0, -8.5, 0, 0, -7.1, 3.8, 0, 0, 7.8, \\ 0, 0]$
\end{itemize}
while its biases are:
\begin{itemize}
    \item $b_1 = 5.9$
    \item $b_2 = 0$
\end{itemize}

\subsubsection{Interpretation}
The tree obtained for this environment is quite simple.
In fact, it uses only two conditions and does not even make use of all the outputs from the output space.

Analyzing the first condition, we observed that, with a small loss per in performance, we can neglect most of the terms, obtaining the simplified condition:
\[7.7\omega^t-8\omega^{l_r}-0.9\omega^{f_r}-6.8\omega^{th_l}+3.3\omega^{f_l} < 5.9,\]
where $t$ stands for torso, $l_r$ for the right leg, $f_r$ for the right foot, $th_l$ for the left thigh, and $f_l$ for the left foot.
Since this condition only includes angular velocities, we hypothesize that this condition tries to understand whether it is safer for a humanoid to make a step or balance itself.
In fact, in one case (i.e., when the condition evaluates to true), it checks what thigh to move, while in the other case (i.e., when it evaluates to false), it moves the right leg, possibly to balance the robot.

Regarding the second condition, numerical experiments showed that the condition can be approximately reduced to:
\[-8.5\theta^{f_l}-7.1\omega^t+7.8\omega^{th_l} < 0,\]
which can be rewritten as
\[\theta^{f_l} > c_1\omega^t + c_2\omega^{th_l}\]

It can be thus interpreted as follows.
If the angle of the left foot is larger than the weighted sum of the angular velocities of the torso and the left thigh, then apply a full-force, positive torque to the left thigh.
Otherwise, apply a full-force torque with the opposite sign to the right thigh.
We hypothesize that this condition tries to understand whether it is more convenient to ``finish'' a step or to raise one of the legs to raise the foot, i.e., it is responsible for making steps, while the first condition is responsible for balancing the robot.

\begin{figure}
    \centering
    \resizebox{0.9\columnwidth}{!}{
    \begin{tikzpicture}
        \node[condition, ] (root) {$\mathbf{w}_1^T\mathbf{x} < b_1$};
        \node[oblique=root/rootl/2/left, xshift=0cm]{};
        \node[leaf, below left = of rootl] (rootll){[1, 0, 0, 0, 0, 0]};
        \node[leaf, below right = of rootl] (rootlr){[0, 0, 0, -1, 0, 0]};
        \draw (rootl) -| node[midway, above]{T}(rootll);
        \draw (rootl) -| node[midway, above]{F}(rootlr);
        \node[leaf, below right = of root] (rootr){[0, 1, 0, 0, 0, 0]};
    \draw (root) -| node[midway, above]{T}(rootl);
    \draw (root) -| node[midway, above]{F}(rootr);

    \end{tikzpicture}
    }
    \caption{Best DT found for the Walker2d-v3 environment.}
    \label{fig:best_w2d}
\end{figure}

\section{Simulation of an episode on selected environments}
In this section, we will showcase some steps of the environment, with the corresponding action, to practically show the advantage of using transparent models for debugging.

\begin{figure*}[!ht]
 \centering
 \begin{subfigure}{0.45\textwidth}
 \includegraphics[width=0.8\columnwidth]{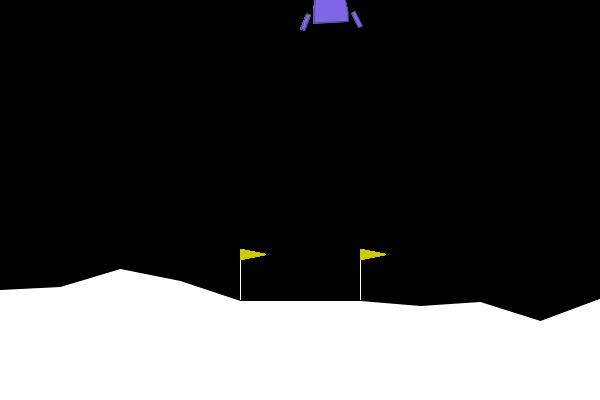}
 \end{subfigure}
 \begin{subfigure}{0.45\textwidth}
 \resizebox{0.8\columnwidth}{!}{
 \begin{tikzpicture}
 \node[condition, fill=blue!20!white] (root) {$\mathbf{w}_1^T\mathbf{x} < b_1$};
 \node[oblique=root/rootl/2/left, fill=blue!20!white] {};
 \node[oblique=rootl/rootll/3/left, xshift=-1cm, fill=blue!20!white] {};
 \node[leaf, below left = of rootll] (rootlll) {2};
 \node[leaf, below right = of rootll, fill=blue!20!white] (rootllr) {1};
 \draw (rootll) -| node[midway, above] {T} (rootlll);
 \draw (rootll) -| node[midway, above] {F} (rootllr);
 \node[oblique=rootl/rootlrr/4/right, xshift=1cm] {};
 \node[oblique=rootlrr/rootlrrl/5/left, yshift=-0.5cm] {};
 \node[leaf, below left = of rootlrrl] (rootlrrll) {3};
 \node[leaf, below right = of rootlrrl] (rootlrrlr) {2};
 \draw (rootlrrl) -| node[midway, above] {T} (rootlrrll);
 \draw (rootlrrl) -| node[midway, above] {F} (rootlrrlr);
 \node[leaf, below right = of rootlrr] (rootlrrr) {0};
 \draw (rootlrr) -| node[midway, above] {T} (rootlrrl);
 \draw (rootlrr) -| node[midway, above] {F} (rootlrrr);
 \draw (rootl) -| node[midway, above] {T} (rootll);
 \draw (rootl) -| node[midway, above] {F} (rootlrr);
 \node[leaf, below right = of root] (rootr) {2};
 \draw (root) -| node[midway, above] {T} (rootl);
 \draw (root) -| node[midway, above] {F} (rootr);
 \end{tikzpicture}
 }
 \end{subfigure}
 \caption{First time step of interest. The agent tries to reduce the velocity on the horizontal axis (towards the right) by pushing the lander to the left.}
 \label{fig:cb_int_t1}
\end{figure*}

\begin{figure*}[!ht]
 \centering
 \begin{subfigure}{0.45\textwidth}
 \includegraphics[width=0.8\columnwidth]{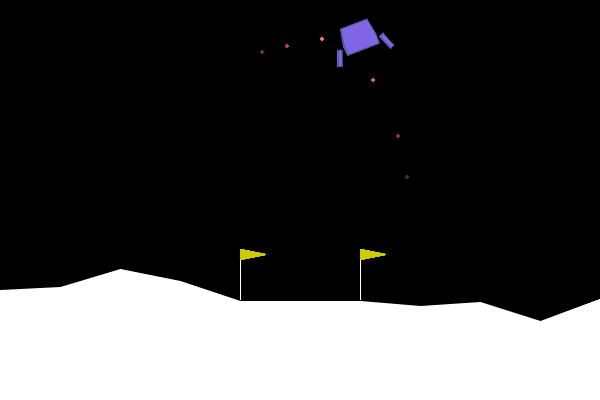}
 \end{subfigure}
 \begin{subfigure}{0.45\textwidth}
 \resizebox{0.8\columnwidth}{!}{
 \begin{tikzpicture}
 \node[condition, fill=blue!20!white] (root) {$\mathbf{w}_1^T\mathbf{x} < b_1$};
 \node[oblique=root/rootl/2/left, fill=blue!20!white] {};
 \node[oblique=rootl/rootll/3/left, xshift=-1cm, fill=blue!20!white] {};
 \node[leaf, below left = of rootll, fill=blue!20!white] (rootlll) {2};
 \node[leaf, below right = of rootll] (rootllr) {1};
 \draw (rootll) -| node[midway, above] {T} (rootlll);
 \draw (rootll) -| node[midway, above] {F} (rootllr);
 \node[oblique=rootl/rootlrr/4/right, xshift=1cm] {};
 \node[oblique=rootlrr/rootlrrl/5/left, yshift=-0.5cm] {};
 \node[leaf, below left = of rootlrrl] (rootlrrll) {3};
 \node[leaf, below right = of rootlrrl] (rootlrrlr) {2};
 \draw (rootlrrl) -| node[midway, above] {T} (rootlrrll);
 \draw (rootlrrl) -| node[midway, above] {F} (rootlrrlr);
 \node[leaf, below right = of rootlrr] (rootlrrr) {0};
 \draw (rootlrr) -| node[midway, above] {T} (rootlrrl);
 \draw (rootlrr) -| node[midway, above] {F} (rootlrrr);
 \draw (rootl) -| node[midway, above] {T} (rootll);
 \draw (rootl) -| node[midway, above] {F} (rootlrr);
 \node[leaf, below right = of root] (rootr) {2};
 \draw (root) -| node[midway, above] {T} (rootl);
 \draw (root) -| node[midway, above] {F} (rootr);
 \end{tikzpicture}
 }
 \end{subfigure}
 \caption{Second time step of interest. The agent fires the main engine to slow down the descent. The position of the lander is slightly tilted due to the previous acceleration towards the left-hand side.}
 \label{fig:cb_int_t2}
\end{figure*}

\begin{figure*}[!ht]
 \centering
 \begin{subfigure}{0.45\textwidth}
 \includegraphics[width=0.8\columnwidth]{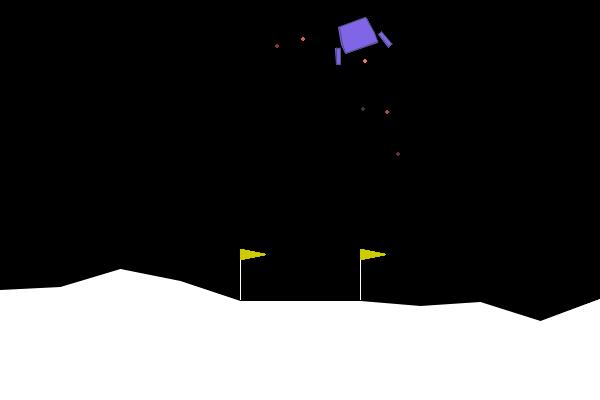}
 \end{subfigure}
 \begin{subfigure}{0.45\textwidth}
 \resizebox{0.8\columnwidth}{!}{
 \begin{tikzpicture}
 \node[condition, fill=blue!20!white] (root) {$\mathbf{w}_1^T\mathbf{x} < b_1$};
 \node[oblique=root/rootl/2/left, fill=blue!20!white] {};
 \node[oblique=rootl/rootll/3/left, xshift=-1cm] {};
 \node[leaf, below left = of rootll] (rootlll) {2};
 \node[leaf, below right = of rootll] (rootllr) {1};
 \draw (rootll) -| node[midway, above] {T} (rootlll);
 \draw (rootll) -| node[midway, above] {F} (rootllr);
 \node[oblique=rootl/rootlrr/4/right, xshift=1cm, fill=blue!20!white] {};
 \node[oblique=rootlrr/rootlrrl/5/left, yshift=-0.5cm, fill=blue!20!white] {};
 \node[leaf, below left = of rootlrrl, fill=blue!20!white] (rootlrrll) {3};
 \node[leaf, below right = of rootlrrl] (rootlrrlr) {2};
 \draw (rootlrrl) -| node[midway, above] {T} (rootlrrll);
 \draw (rootlrrl) -| node[midway, above] {F} (rootlrrlr);
 \node[leaf, below right = of rootlrr] (rootlrrr) {0};
 \draw (rootlrr) -| node[midway, above] {T} (rootlrrl);
 \draw (rootlrr) -| node[midway, above] {F} (rootlrrr);
 \draw (rootl) -| node[midway, above] {T} (rootll);
 \draw (rootl) -| node[midway, above] {F} (rootlrr);
 \node[leaf, below right = of root] (rootr) {2};
 \draw (root) -| node[midway, above] {T} (rootl);
 \draw (root) -| node[midway, above] {F} (rootr);
 \end{tikzpicture}
 }
 \end{subfigure}
 \caption{Third time step of interest. The agent pushes the lander to the right to restore the correct angle.}
 \label{fig:cb_int_t3}
\end{figure*}

\begin{figure*}[!ht]
 \centering
 \begin{subfigure}{0.45\textwidth}
 \includegraphics[width=0.8\columnwidth]{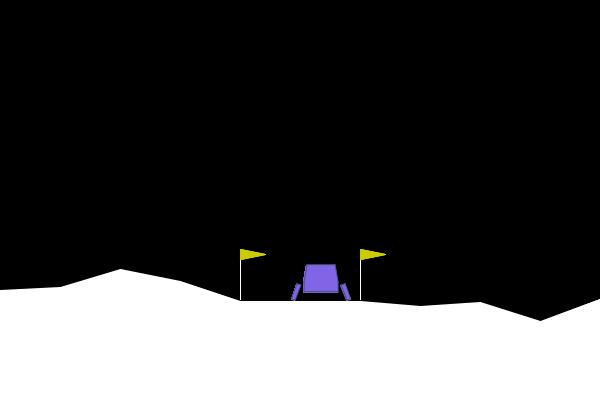}
 \end{subfigure}
 \begin{subfigure}{0.45\textwidth}
 \resizebox{0.8\columnwidth}{!}{
 \begin{tikzpicture}
 \node[condition, fill=blue!20!white] (root) {$\mathbf{w}_1^T\mathbf{x} < b_1$};
 \node[oblique=root/rootl/2/left, fill=blue!20!white] {};
 \node[oblique=rootl/rootll/3/left, xshift=-1cm] {};
 \node[leaf, below left = of rootll] (rootlll) {2};
 \node[leaf, below right = of rootll] (rootllr) {1};
 \draw (rootll) -| node[midway, above] {T} (rootlll);
 \draw (rootll) -| node[midway, above] {F} (rootllr);
 \node[oblique=rootl/rootlrr/4/right, xshift=1cm, fill=blue!20!white] {};
 \node[oblique=rootlrr/rootlrrl/5/left, yshift=-0.5cm] {};
 \node[leaf, below left = of rootlrrl] (rootlrrll) {3};
 \node[leaf, below right = of rootlrrl] (rootlrrlr) {2};
 \draw (rootlrrl) -| node[midway, above] {T} (rootlrrll);
 \draw (rootlrrl) -| node[midway, above] {F} (rootlrrlr);
 \node[leaf, below right = of rootlrr, fill=blue!20!white] (rootlrrr) {0};
 \draw (rootlrr) -| node[midway, above] {T} (rootlrrl);
 \draw (rootlrr) -| node[midway, above] {F} (rootlrrr);
 \draw (rootl) -| node[midway, above] {T} (rootll);
 \draw (rootl) -| node[midway, above] {F} (rootlrr);
 \node[leaf, below right = of root] (rootr) {2};
 \draw (root) -| node[midway, above] {T} (rootl);
 \draw (root) -| node[midway, above] {F} (rootr);
 \end{tikzpicture}
 }
 \end{subfigure}
 \caption{Fourth time step of interest. The agent does not fire any engine as the lander successfully landed.}
 \label{fig:cb_int_t4}
\end{figure*}

See \Cref{fig:cb_int_t1,fig:cb_int_t2,fig:cb_int_t3,fig:cb_int_t4} for a graphical visualization and a textual description of the scenarios in LunarLander-v2.

\begin{figure*}[!ht]
 \centering
 \begin{subfigure}{0.45\textwidth}
 \includegraphics[width=0.8\columnwidth]{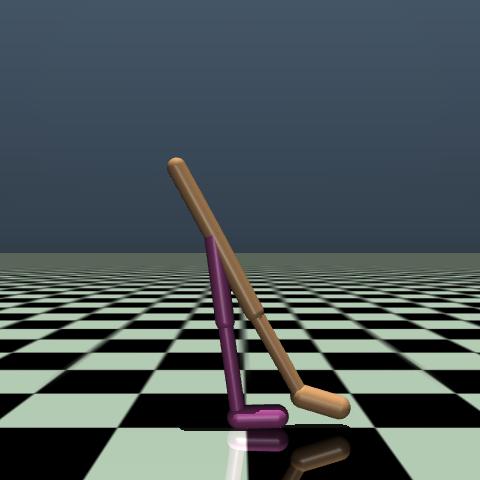}
 \end{subfigure}
 \begin{subfigure}{0.45\textwidth}
 \resizebox{0.9\columnwidth}{!}{
    \begin{tikzpicture}
        \node[condition, fill=blue!20!white] (root) {$\mathbf{w}_1^T\mathbf{x} < b_1$};
        \node[oblique=root/rootl/2/left, xshift=0cm, fill=blue!20!white]{};
        \node[leaf, below left = of rootl, fill=blue!20!white] (rootll){[1, 0, 0, 0, 0, 0]};
        \node[leaf, below right = of rootl] (rootlr){[0, 0, 0, -1, 0, 0]};
        \draw (rootl) -| node[midway, above]{T}(rootll);
        \draw (rootl) -| node[midway, above]{F}(rootlr);
        \node[leaf, below right = of root] (rootr){[0, 1, 0, 0, 0, 0]};
    \draw (root) -| node[midway, above]{T}(rootl);
    \draw (root) -| node[midway, above]{F}(rootr);

    \end{tikzpicture}
    }
 \end{subfigure}
 \caption{First time step of interest. The agent applies a torque of $1 N\cdot m$ to the right thigh to lift the right foot.}
 \label{fig:cb_int_t1_w2d}
\end{figure*}

\begin{figure*}[!ht]
 \centering
 \begin{subfigure}{0.45\textwidth}
 \includegraphics[width=0.8\columnwidth]{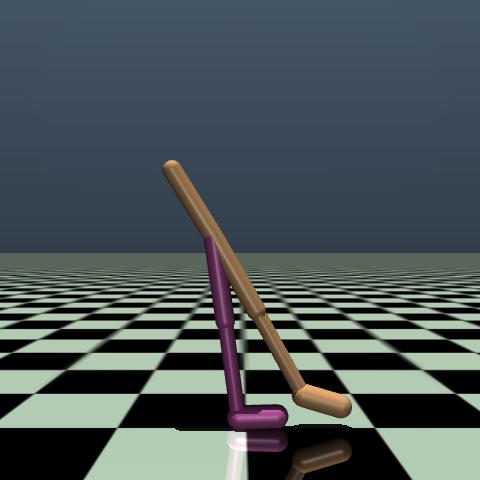}
 \end{subfigure}
 \begin{subfigure}{0.45\textwidth}
 \resizebox{0.9\columnwidth}{!}{
    \begin{tikzpicture}
        \node[condition, fill=blue!20!white] (root) {$\mathbf{w}_1^T\mathbf{x} < b_1$};
        \node[oblique=root/rootl/2/left, xshift=0cm]{};
        \node[leaf, below left = of rootl] (rootll){[1, 0, 0, 0, 0, 0]};
        \node[leaf, below right = of rootl] (rootlr){[0, 0, 0, -1, 0, 0]};
        \draw (rootl) -| node[midway, above]{T}(rootll);
        \draw (rootl) -| node[midway, above]{F}(rootlr);
        \node[leaf, below right = of root, fill=blue!20!white] (rootr){[0, 1, 0, 0, 0, 0]};
    \draw (root) -| node[midway, above]{T}(rootl);
    \draw (root) -| node[midway, above]{F}(rootr);
    \end{tikzpicture}
    }
 \end{subfigure}
 \caption{Second time step of interest. The agent applies a torque of $1 N\cdot m$ to the right leg to bend the right knee.}
 \label{fig:cb_int_t2_w2d}
\end{figure*}

\begin{figure*}[!ht]
 \centering
 \begin{subfigure}{0.45\textwidth}
 \includegraphics[width=0.8\columnwidth]{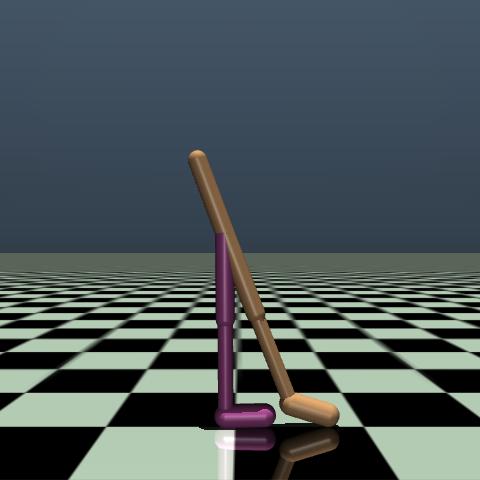}
 \end{subfigure}
 \begin{subfigure}{0.45\textwidth}
 \resizebox{0.9\columnwidth}{!}{
    \begin{tikzpicture}
        \node[condition, fill=blue!20!white] (root) {$\mathbf{w}_1^T\mathbf{x} < b_1$};
        \node[oblique=root/rootl/2/left, xshift=0cm, fill=blue!20!white]{};
        \node[leaf, below left = of rootl] (rootll){[1, 0, 0, 0, 0, 0]};
        \node[leaf, below right = of rootl, fill=blue!20!white] (rootlr){[0, 0, 0, -1, 0, 0]};
        \draw (rootl) -| node[midway, above]{T}(rootll);
        \draw (rootl) -| node[midway, above]{F}(rootlr);
        \node[leaf, below right = of root] (rootr){[0, 1, 0, 0, 0, 0]};
    \draw (root) -| node[midway, above]{T}(rootl);
    \draw (root) -| node[midway, above]{F}(rootr);

    \end{tikzpicture}
    }
 \end{subfigure}
 \caption{Third time step of interest. The agent applies a torque of $-1 N\cdot m$ to the left thigh to place the right foot.}
 \label{fig:cb_int_t3_w2d}
\end{figure*}

On the other hand, \Cref{fig:cb_int_t1_w2d,fig:cb_int_t2_w2d,fig:cb_int_t3_w2d} show timesteps from the Walker2d-v3 environment.

\section{Comparison with ELDT}
In this section, given the similarities between ELDT \cite{custode2023evolutionary} and SIRL, we compare the two approaches more closely.
While the two methods share many similarities, it is important to note that ELDT is a non-social approach, while SIRL is a social-learning approach.
While this may seem of secondary importance, it actually has important consequences, described in the following.

\paragraph{Computational cost}
Given a population size $p$, $e$ episodes in ELDT, $e_c$ collaborative episodes in SIRL, and $e_i$ individual episodes in SIRL, we have that ELDT has a cost (\textit{per generation}) of $O(p\cdot e)$, while SIRL has a cost of $O(e_c + p\cdot e_i)$.
This means that we can choose $e_c$ and $e_i$ in such a way that $p\cdot e > e_c + p\cdot e_i$, which allows SIRL to be significantly more efficient than ELDT.

\paragraph{Episodes experienced by each individual}
In ELDT, each DT experiences $e$ episodes.
In SIRL, each DT experiences $e_c + e_i$ episodes.
This means that, by appropriately choosing $e_c$ and $e_i$, we can not only reduce the computational cost of the process, but also allow each DT to experience significantly more episodes, allowing for the learning of a better state-action function.

\begin{figure}
 \centering
 \includegraphics[width=0.9\columnwidth]{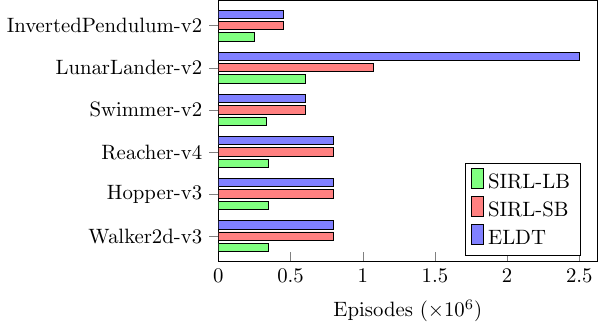}
 \caption{Number of episodes used by each of the algorithms under comparison.}
 \label{fig:budget2}
\end{figure}

\paragraph{Experimental comparison}
In \Cref{fig:dts_trends}, we can see a comparison between ELDT and SIRL based on the best score obtained for each generation.
Here, we test two different setups for SIRL: SB (which stands for ``Same Budget as ELDT'') and LB (``Lower Budget than ELDT''), where LB roughly uses half of the episodes used by ELDT (see \Cref{fig:budget2}).
Note that, in LunarLander, the cost of ELDT was prohibitive, so also SIRL-SB has a lower budget in this case.

\begin{figure*}[!t]
 \centering
 \includegraphics[width=0.8\textwidth]{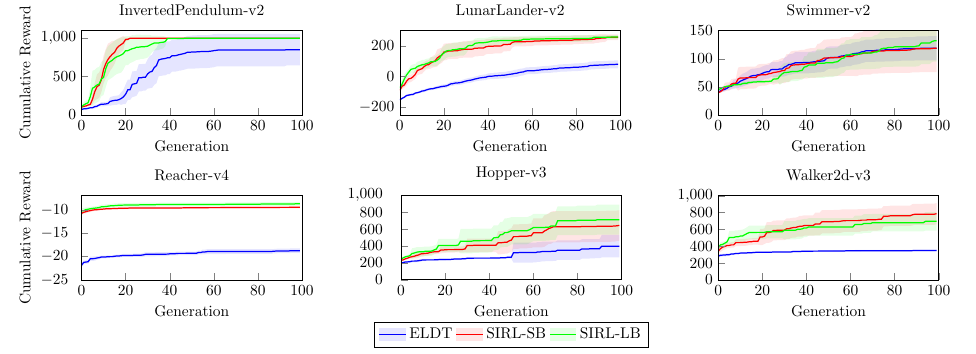}
 \caption{Scores (the higher, the better) obtained by the best agent, at each iteration of the outer loop (i.e., a generation), for the population-based training of interpretable agents. The solid line represents the mean value, while the shaded area represents the $95\%$ CI.
 }
 \label{fig:dts_trends2}
\end{figure*}

From \Cref{fig:dts_trends2}, we can easily observe that SIRL often achieves better convergence than ELDT, even with the low-budget setup, confirming that the social learning scenario actually improves the effectiveness of our algorithm. This, in our opinion, is due to the fact that each agent experiences a significantly higher number of episodes, which enables better learning.

In \Cref{fig:dts_boxplot}, instead, we compare the final scores for the agents evolved through ELDT, SIRL-SB, and SIRL-LB.
It is easy to observe that SIRL always achieves better performance w.r.t. ELDT, even in the LB case.
However, in Walker2d-v3, we observe that SIRL-LB has a lower score w.r.t. SIRL-SB, which indicates that, for challenging environments, a high budget can lead to significantly better performance.

\paragraph{Statistical tests}
To validate the statistical significance of our results, we perform a one-way ANOVA test on the results, using the post-hoc Tukey HSD.
In \Cref{tab:anova} we show the results of the pairwise comparisons between the tested methods.
We use a confidence threshold of $\alpha = 0.05$.

We observe that, in $4$ tasks over $6$, we can reject the null hypothesis between at least one of the SIRL methods and ELDT.
This indicates that SIRL, in the worst case, achieves comparable performance with respect to ELDT while using a substantially smaller number of episodes.

\begin{table}[!ht]
 \centering
 \caption{Results of the pairwise comparisons using the one-way ANOVA test with post-hoc Tukey HSD. Bold values indicate statistically significant results.}
 \label{tab:anova}
 \resizebox{0.9\columnwidth}{!}{
 \begin{tabular}{l l c} \toprule
 \textbf{Environment} & \textbf{Pairwise comparison} & \textbf{p-value} \\ \midrule
 \multirow{3}{*}{InvertedPendulum-v2} & ELDT vs SIRL-SB & 0.198 \\
 & ELDT vs SIRL-LB & 0.245 \\
 & SIRL-SB vs SIRL-LB & 0.991 \\ \midrule
 \multirow{3}{*}{LunarLander-v2} & ELDT vs SIRL-SB & \textbf{0.031} \\
 & ELDT vs SIRL-LB & \textbf{0.014} \\
 & SIRL-SB vs SIRL-LB & 0.090 \\ \midrule
 \multirow{3}{*}{Swimmer-v2} & ELDT vs SIRL-SB & 0.988 \\
 & ELDT vs SIRL-LB & 0.989 \\
 & SIRL-SB vs SIRL-LB & 0.954 \\ \bottomrule
 \multirow{3}{*}{Reacher-v4} & ELDT vs SIRL-SB & \textbf{0.000} \\
 & ELDT vs SIRL-LB & \textbf{0.002} \\
 & SIRL-SB vs SIRL-LB & 0.830 \\ \bottomrule
 \multirow{3}{*}{Hopper-v3} & ELDT vs SIRL-SB & 0.363 \\
 & ELDT vs SIRL-LB & \textbf{0.013} \\
 & SIRL-SB vs SIRL-LB & 0.234 \\ \bottomrule
 \multirow{3}{*}{Walker2d-v3} & ELDT vs SIRL-SB & \textbf{0.001} \\
 & ELDT vs SIRL-LB & 0.292 \\
 & SIRL-SB vs SIRL-LB & \textbf{0.041} \\ \bottomrule
 \end{tabular}
 }
\end{table}

\begin{figure*}
 \centering
 \includegraphics[width=0.8\textwidth]{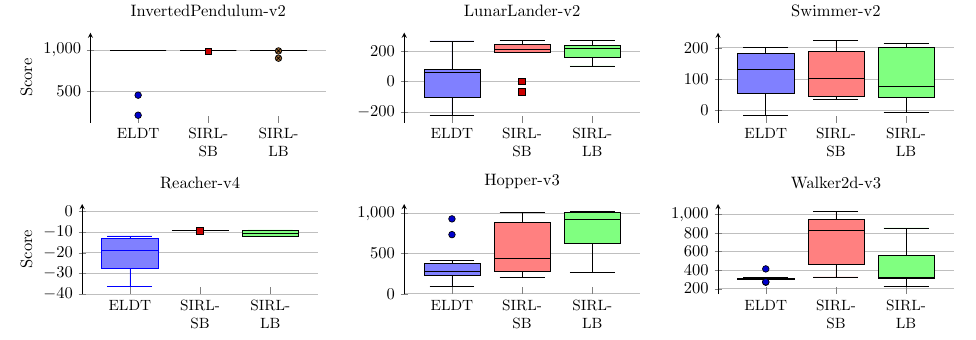}
 \caption{Scores (the higher, the better) obtained by the best agents found by each tested method in $10$ runs, tested on $100$ unseen episodes.}
 \label{fig:dts_boxplot2}
\end{figure*}

\end{document}